\documentclass[manuscript]{acmart}
\AtBeginDocument{%
	\providecommand\BibTeX{{%
			\normalfont B\kern-0.5em{\scshape i\kern-0.25em b}\kern-0.8em\TeX}}}

\setcopyright{acmcopyright}
\copyrightyear{2022}
\acmYear{2022}
\acmDOI{10.1145/1122445.1122456}

\acmJournal{csur}
\acmVolume{37}
\acmNumber{4}
\acmArticle{111}
\acmMonth{8}

\usepackage{url}
\usepackage{xspace,mfirstuc,tabulary}
\usepackage{hhline}
\usepackage{booktabs}
\usepackage{multirow}
\usepackage{subfigure}
\usepackage{amsmath}
\usepackage{graphicx}
\usepackage[figuresright]{rotating}
\usepackage{makecell}
\usepackage{diagbox}
\usepackage{bbding}

\newcommand{\secref}[1]{Section \ref{#1}}
\newcommand{\figref}[1]{Figure \ref{#1}}

\newcommand{\cut}[1]{}
\newcommand{\tabincell}[2]{\begin{tabular}{@{}#1@{}}#2\end{tabular}}

\begin{document}
	
\title{Taxonomy of Abstractive Dialogue Summarization: Scenarios, Approaches and Future Directions}


\author{Qi Jia}
\affiliation{%
	\institution{Shanghai Jiao Tong University}
	\streetaddress{800 Dongchuan Road}
	\city{Shanghai}
	\country{China}
	\postcode{200240}}
\email{Jia\_qi@sjtu.edu.cn}

\author{Yizhu Liu}
\affiliation{%
	\institution{Meituan}
	\city{Shanghai}
	\country{China}
	\postcode{200093}}
\email{liuyizhu@meituan.com}

\author{Siyu Ren}
\affiliation{%
	\institution{Shanghai Jiao Tong University}
	\streetaddress{800 Dongchuan Road}
	\city{Shanghai}
	\country{China}
	\postcode{200240}}
\email{roy0702@sjtu.edu.cn}

\author{Kenny Q. Zhu}
\authornote{Corresponding author.}
\affiliation{%
	\institution{University of Texas at Arlington}
	\streetaddress{500 UTA Blvd}
	\city{Arlington}
	\state{TX}
	\country{United States}
	\postcode{76010}}
\email{kenny.zhu@uta.edu}

\renewcommand{\shortauthors}{Jia et al.}

\begin{abstract}
	Abstractive dialogue summarization generates a concise and fluent summary covering the salient information in a dialogue among two or more interlocutors.
	It has attracted significant attention in recent years based on the massive emergence of social communication platforms and an urgent requirement for efficient dialogue information understanding and digestion.
	Different from news or articles in traditional document summarization, 
	dialogues bring unique characteristics and additional challenges, including different language styles and formats, scattered information, flexible discourse structures, and unclear topic boundaries.
	This survey provides a comprehensive investigation of existing work for abstractive dialogue summarization from scenarios, approaches to evaluations.
	It categorizes the task into two broad categories according to the type
	of input dialogues, i.e., open-domain and task-oriented, and
	presents a taxonomy of existing techniques in three directions, 
	namely, injecting dialogue features, designing auxiliary training tasks and 
	using additional data.
	A list of datasets under different scenarios and widely-accepted evaluation metrics 
	are summarized for completeness. 
	After that, the trends of scenarios and techniques are summarized, together with deep insights into correlations between extensively exploited features and different scenarios.
	Based on these analyses, we recommend future directions, including more controlled and complicated scenarios, technical innovations and comparisons, publicly available datasets in special domains, etc.
\end{abstract}



\maketitle

\section{Introduction}

Abstractive text summarization aims at generating a concise summary output covering key points given the source input.
Prior studies mainly focus on narrative text inputs such as news stories 
, including CNN/DM~\cite{hermann2015teaching} and XSum~\cite{narayan2018don}, 
and other publications, including PubMed and 
arXiv~\cite{cohan2018discourse}, and have achieved remarkable success.  
As a natural way of communication, dialogues have attracted 
increasing attention in recent years. 
With the rapid growth of real-time messaging, consultation 
forums, and online meetings,
information explosion in the form of dialogues calls for more efficient 
ways of searching and digesting dialogues.

Dialogue summarization targets summarizing salient information in a third party's view given utterances among two or 
more interlocutors. 
This task is not only helpful in providing a quick context to new participants of a conversation, 
but can also help people grasp the central ideas or search for key contents in the conversation, which promotes efficiency and
productivity. 
It is first proposed as meeting summarization by \citet{carletta2005ami} and \citet{janin2003icsi} and generally
covers a number of scenarios, such as daily chat~\cite{gliwa2019samsum,chen2021dialsumm}, 
medical consultation~\cite{joshi2020dr}, customer service~\cite{zou2021topic}, etc.
Different from document summarization where inputs are narrative texts from a third party, inputs for dialogue summarization are uttered by multiple parties in the first person. 
Dialogues are not only abundant with informal expressions and elliptical utterances~\cite{zhang2020filling,liu2020incomplete},
but also full of question answerings, and repeated confirmations to reach a consensus among speakers.
The inherent semantic flows are complicatedly reflected by vague topic boundaries~\cite{takanobu2018weakly} and
interleaved inter-utterance dependencies~\cite{afantenos2015discourse}. 
In a word, the information in dialogues is sparse and less structured, and the utterances are highly content-dependent, raising the difficulty for dialogue summarization.


Based on these characteristics, abstractive dialogue summarization generating fluent summaries is preferred by humans instead of the extractive one that extracts utterances.
The earliest efforts approached this by transforming dialogues into word graphs and selecting the suitable paths in the graph as summary sentences by complicated rules~\cite{banerjee2015generating,shang2018unsupervised}.
Template-based approaches~\cite{OyaMCN14,singla2017spoken} were also adopted, which collect templates from human-written summaries and generate abstractive summaries by selecting suitable words from the dialogue to fill in the blank. However, their generated summaries lack fluency and diversity thus are far from practical use.  
Later, neural encoder-decoder models showed up. They projected the input into dense semantic representations and summaries with novel words were generated by sampling from the vocabulary list step-by-step until a special token representing the end was emitted. Abstractive text summarization has achieved remarkable progress based on these models tracing back from 
non-pretrained ones such as PGN~\cite{see2017get}, Fast-Abs~\cite{chen2018fast} and HRED~\cite{serban2016building}, to pretrained ones including 
BART~\cite{lewis2020bart} and Pegasus~\cite{zhang2020pegasus}. 
At the same time, techniques for dialogue context modeling have also evolved significantly with neural models in dialogue-related researches, such as dialogue reading comprehension~\cite{sun2019dream}, response selection~\cite{xu2020learning} and dialogue information extraction~\cite{yu2020dialogue}.
The rapid growth of the two areas above paves the way for a recent revival of research in abstractive dialogue summarization.



Dozens of papers have been published in the area of dialogue summarization in recent years. 
Notably, a number of technical papers have dug into various dialogue features and datasets under different scenarios. 
It is time to look at what has been achieved, find potential omissions and provide a basis for future work.
However, there is no comprehensive review of this field, except for Feng et al.'s recent survey~\cite{feng2021survey}. 
Different from their paper which focuses on datasets and benchmarks targetting only a few applications, 
our survey aims at providing a thorough account of abstractive dialogue summarization, containing taxonomies of task formulations with different scenarios, various techniques, and evaluations covering different metrics and $29$ datasets.
This survey not only serves as a review of existing work and points out future directions for research but also can be a useful look-up manual 
for engineers when solving problems. We also hope this survey could serve as a milestone for dialogue summarization approaches mainly before the emergence of large language models (LLM), such as LLaMa~\cite{touvron2023llama} and ChatGPT~\cite{openai2022}, and bring inspirations for developing new techniques with LLMs.


The remainder of this review is structured as follows. \secref{sec:task} is the problem formulation, providing a formal task definition, 
unique characteristics compared to document summarization and hierarchical classification of existing application scenarios.
\secref{sec:approach} to \secref{sec:useadddata} presents a comprehensive taxonomy of dialogue summarization approaches in which current dialogue 
summarization techniques are mainly based on tested document summarization models and can be divided into 
three directions, including (1) injecting pre-processed features~(\secref{sec:feature}), (2) designing self-supervised tasks~(\secref{sec:designselftasks}), and 
(3) using additional data~(\secref{sec:useadddata}).
A collection of proposed datasets and evaluation metrics are in \secref{sec:evaluation}.
Based on the highly related papers, 
we offer deep insights on correlations between techniques and scenarios in Section~\ref{sec:observations}. 
We further suggest several future directions, including more controlled and complicated 
scenarios, technical innovations and feature comparisons, open-source datasets in special domains, 
and benchmarks and methods for evaluation in \secref{sec:future}.

\section{Problem Formulation}
\label{sec:task}

In this section, we formally define the abstractive dialogue summarization
task with mathematical notations. We highlight the characteristics of this task by contrasting it with the well-studied document summarization
problem. Finally, we present a hierarchical classification of application scenarios, demonstrating the practicality of this task.

\subsection{Task Definition}\label{sec:taskdefinition}
A dialogue can be formalized as a sequence of $T$ chronologically ordered turns:
\begin{equation}
	D = \{U_1, U_2, ..., U_T\}
	\label{eq:dialogue}
\end{equation}
Each turn $U_t$ generally consists of a speaker/role $s_t$ and corresponding utterance $u_t = \{w_i^t|_{i=1}^{l_t}\}$. $w_i^t$ represents the $i$-th token\footnote{To construct input for neural models, tokenizers are used to tokenize utterances into tokens in the vocabulary. Rare words may result in multiple tokens by algorithms such as Byte-Pair-Encoding. We do not strictly distinguish words and tokens in this survey.} in the $t$-th utterance, $l_t$ is the length of $u_t$.

Dialogue summarization aims at generating a short but informative 
summary $Y=\{y_1,y_2,...,y_n\}$ for $D$, where $n$ is 
the number of summary tokens. $Y$ represents the reference summary 
and $\hat{Y}$ represents the generated summary.

\subsection{Comparisons to Document Summarization}\label{sec:divergence}

Dialogue summarization is different from document summarization in various 
aspects, including language style and format, information density, 
discourse structure, and topic boundaries.

\begin{figure}[ht]
	\centering
	\includegraphics[scale=0.5]{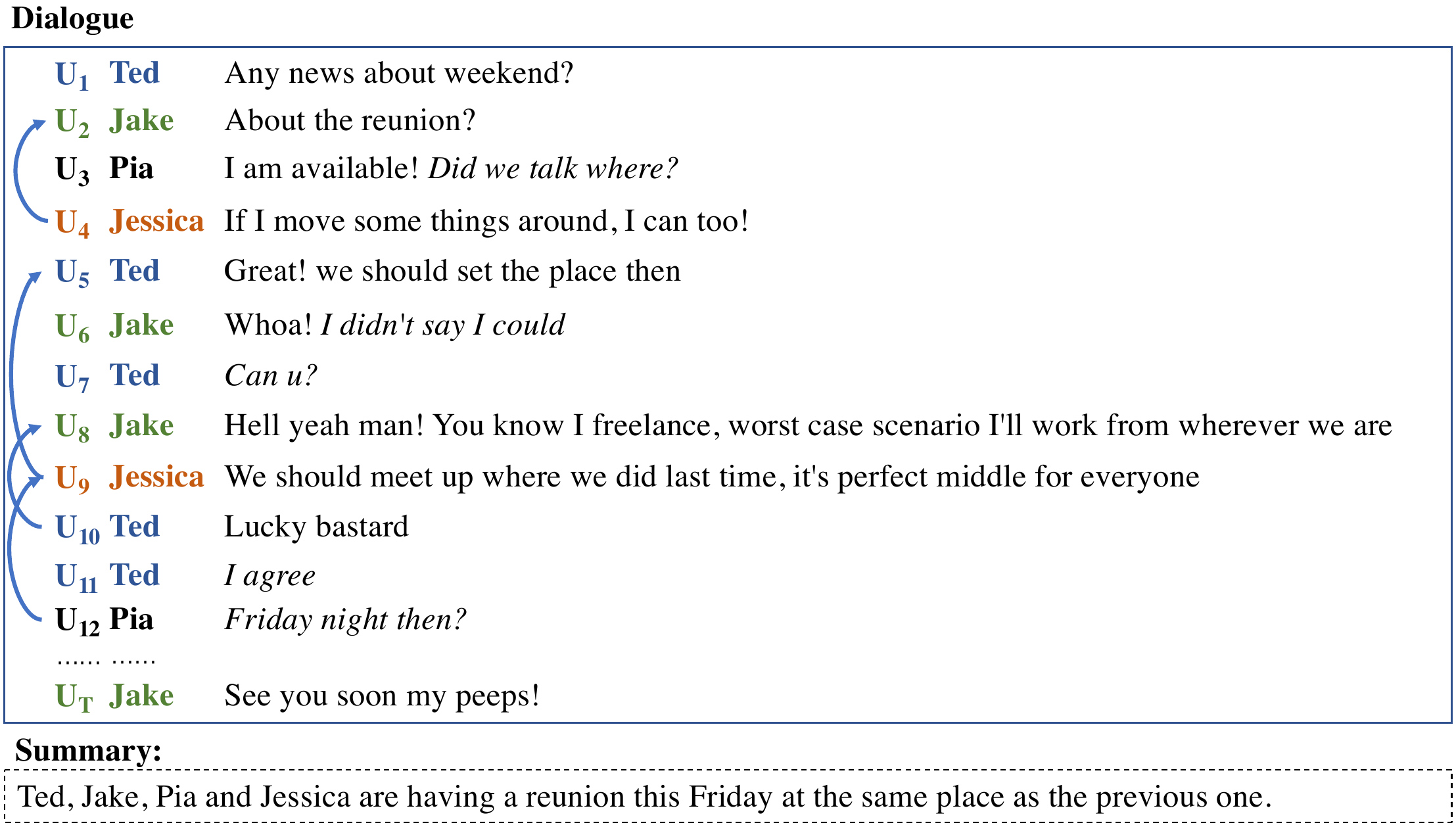}
	\caption{An example multi-party dialogue and its summary. The arrows represent unsequential dependencies between utterances. Elliptical sentences are in italic.}
	\label{fig:example}
\end{figure}

\textbf{Word Level - Language Style and Format:} 
Documents in previous well-researched summarization tasks are written from the third point of view, while dialogues consist of utterances expressed by different speakers in first person. Informal and colloquial expressions are common especially for recorded dialogues from speech, such as ``Whoa'' in $U_6$ and ``u'' representing ``you'' in $U_7$ from Figure~\ref{fig:example}.
Pronouns are frequently used to refer to events or persons mentioned in the dialogue history. Around 72\% of mentions in the conversation are {anaphoras} 
as stated in \citet{bai2021joint}. Meanwhile, the performance of coreference resolution models trained on normal text drops dramatically on dialogues~\cite{liu2021coreference}. It manifests the existence of language style differences between documents and dialogues, leading to difficulties in understanding the mappings between speakers and events in dialogues.

\textbf{Sentence/Utterance Level - Information Density:}
Document sentences are more self-contained with complete SVO (subject-verb-object) structures, while elliptical utterances are ubiquitous in dialogues, including $U_3$, $U_6$, $U_7$, $U_{11}$ and $U_{12}$.
Besides, the long dialogue can be summarized into a single summary sentence for the example in Figure~\ref{fig:example} as a result of back-and-forth questions and confirmations among speakers for communication purposes.
Question answerings, acknowledgments, and comments~\cite{asher2016discourse} are frequent discourse relations among utterances to narrow down speakers' information gaps and reach agreements.
In this way, dialogue utterances are highly content-dependent, and the information is scattered~\cite{zhang2021exploratory}, raising the difficulties for generating integral contents.

\textbf{Inter-sentence/utterance Level - Discourse structure:}
Articles tend to be well-structured, such as 
general-to-specific structure or deductive order. 
For example, the most important information 
in news summarization are always at the beginning of the document, resulting in a competitive performance of the simple Lead-$3$ baseline~\cite{nallapati2017, see2017get}. However, it is not the same for dialogue summarization. Both Lead-$3$ and Longest-$3$, i.e. $\{U1, U2, U3\}$ and $\{U4, U8, U9\}$ in Figure~\ref{fig:example}, get poor results in different dialogue scenarios~\cite{gliwa2019samsum,chen2021dialsumm,zhang2021emailsum}.
The dependencies among utterances are interleaved, shown by arrows in Figure~\ref{fig:example}, and discourse relations in dialogues are more flexible,
even with the correction of wrong information~\cite{asher2016discourse}. 
For example, 
Jake refused to be available for the reunion in $U_6$, but later agreed
in $U_8$.  As a result, it is more challenging to reason cross utterances for 
dialogue summarization than document summarization.

\textbf{Passage/Session Level - Topic boundaries:} Sentences under the same topic in documents are collected together in a paragraph or a section.
Previous works for extractive~\cite{xiao2019extractive} and abstractive summarization~\cite{cohan2018discourse} both took advantage of such features and made great progress. 
However, a dialogue is a stream of continuous 
utterances without boundaries, even for hours of discussion. The same topic may be discussed repeatedly 
with redundancies and new information, setting up obstacles for content 
selection in dialogue summarization.

To better explain why abstractive approaches are more preferred than extractive ones for dialogues, we list the result of the best rule-based extractive baseline, i.e., Longest-$3$~\cite{gliwa2019samsum}, the oracle extractive result determined by Rouge-L Recall score between each summary sentence and dialogue utterances~\cite{chen2018fast}, and the generation by BART fine-tuned on SAMSum dataset~\cite{gliwa2019samsum} of the simple dialogue in Figure~\ref{fig:example} as follows:
\\
\begin{tabular}{|p{1.5cm}|p{\linewidth-2.3cm}|}
	\hline
	\textbf{Longest-$3$} & Jessica: If I move some things around, I can too! Jake: Hell yeah man! You know I freelance, worst case scenario I'll work from wherever we are Jessica: We should meet up where we did last time, it's perfect middle for everyone.\\
	\hline
	\textbf{Oracle} & Jake: Hell yeah man! You know I freelance, worst case scenario I'll work from wherever we are\\
	\hline
	\textbf{BART}& Ted, Pia, Jessica and Jake are going to meet up on Friday night. \\
	\hline
\end{tabular}
\\
We can see that the readability of generated summaries are poor for Longest-$3$ and Oracle due to the language style and format difference. The compression ratio of Longest-$3$ is apparently low while it still misses the involvement of Ted and Pia as a result of low information density of dialogues. Oracle is concise but much more information is missing. Meanwhile, the fine-tuned BART as an abstractive approach shows the favorable performance. Therefore, abstractive approaches becomes the mainstream in researches on dialogue summarization. In a word, dialogue summarization is an valuable research direction in summarization, where the modeling and understanding of dialogues are challenging compared with document summarization and abstractive approaches are especially preferred.



\subsection{Scenarios for Dialogue Summarization}\label{sec:scenarios}

\begin{figure}
	\centering
	\includegraphics[scale=0.8]{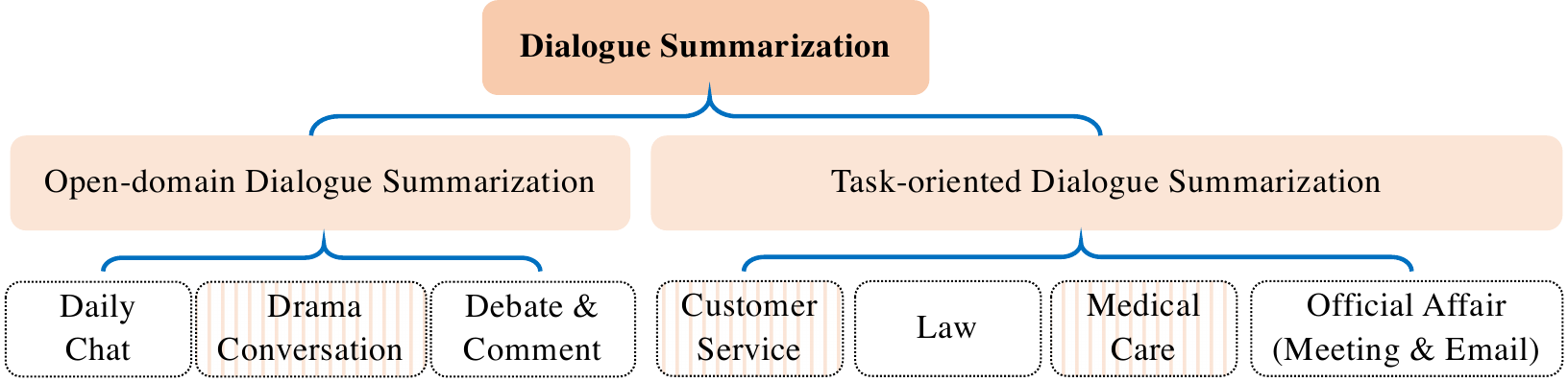}
	\caption{The classification of dialogue summarization tasks with different application scenarios. Datasets proposed for evaluations under each scenario are in Section~\ref{sec:dataset}.}
	\label{fig:scenario}
\end{figure}

Considering the source of dialogues and the purpose of doing summarization,
we divide the application scenarios into two classes: \textbf{open-domain dialogue summarization (ODS)} and \textbf{task-oriented dialogue summarization (TDS)}. This taxonomy is similar to the one of dialogue systems~\cite{gao2020standard,chen2017survey}.
However, one should note that a pre-defined domain ontology for dialogues is 
not necessarily required for TDS, which is different from that in 
task-oriented dialogue systems.
The application scenarios investigated in previous papers 
are classified into these two classes as shown in Figure~\ref{fig:scenario}.

Open-domain dialogue summarization is further divided into daily chat, 
drama conversation, debate \& comment. 
\textbf{Daily chat}~\cite{gliwa2019samsum,chen2021dialsumm} refers to the dialogues happening in our daily lives, 
such as making appointments, discussions between friends, etc. 
\textbf{Drama conversation}~\cite{rameshkumar2020storytelling,zhu2021mediasum,malykh2020sumtitles,chen2021summscreen} represents dialogues from soap operas, 
movies or TV shows, which are dramatized or fabricated with drama scripts 
behind them. Dialogues in these two classes are full of person names 
and events, resulting in narrative summaries about ``who did what''.
\textbf{Debate \& comment}~\cite{misra2015using,fabbri2021convosumm,chowdhury2019cqasumm} focuses more on question answering and 
discussions in online forums and arguments. These dialogues emphasize opinions or solutions to the given subject or questions.

Task-oriented dialogue summarization arises from application scenarios of different domains, which includes but is not limited to customer service, 
law, medical care and official issue.
\textbf{Customer service}~\cite{zou2021topic,feigenblat-etal-2021-tweetsumm-dialog,zhao2021todsum,liu2019automatic,chen2020jddc} refers to conversations between customers and service providers.
Customers start the conversation with their specific intents and agents are required to meet these requirements with the help of their in-domain databases, such as hotel reservations and express information consultation for online shopping. Dialogue summarization for this task is mainly to help service providers quickly go through solutions to users' questions for agent training and service evaluation. 
\textbf{Law}~\cite{fuzw20,duan2019legal,xi2020global} is dialogues related to legal service and 
criminal investigations. Dialogue summarization in this scenario alleviates the recording and summarizing workload 
for law enforcement or legal professionals. 
\textbf{Medical care}~\cite{joshi2020dr,song2020summarizing,song2020summarizing,zhang2021leveraging,liu2019topic} is dialogues between doctors and patients and medical dialogue summarization has some similarity to the research on electronic health records (EHR). Unlike the previous work focusing on mining useful information from EHR~\cite{yadav2018mining}, summarization is to extract useful information from the doctor-patient dialogue and generate an EHR-like or fluent summary for clinical decision-making or online search. It also aims to reduce the burden of domain experts.
\textbf{Official affair}~\cite{carletta2005ami,janin2003icsi,ulrich2008publicly,zhang2021emailsum} is conversations between colleagues for technical or teachers and students for academic issue discussion. They can be either in the format of meetings or e-mails, with summaries covering problems, solutions, and plans.

We compare and contrast ODS and TDS as follows.
\begin{itemize}
\item Dialogues happen between \textbf{two or more speakers} both in ODS and TDS, whereas the \textbf{interpersonal relationship} and \textbf{functional relationship} among speakers are different. Generally, speakers in ODS are friends, neighbors, lovers, family members, and so on. 
They are equal either in the aspect of interpersonal relationships or functional relationships. For example, one can raise a question or answer others' questions in online forums~\cite{fabbri2021convosumm}.
In TDS, speakers have different official roles acting for corresponding responsibilities. For example, plaintiff, defendant, witness and judge in court debates~\cite{duan2019legal}, project manager, marketing expert, user interface designer and industrial designer in official meetings~\cite{carletta2005ami} are corresponding roles.
Among different dialogues, roles are the same and can be played by different speakers and a speaker's role is always unchanged for a service platform.
In a word, TDS pays more attention to functional roles while ODS focuses on speakers.

\item Multiple \textbf{topics} may be covered in the same dialogue session.
Topics in ODS are more diverse than in TDS. The summarization models are expected to deal with unlimited open-domain topics such as chitchat, sales, education, and climate at the same time~\cite{chen2021dialsumm}. 
However, topics in TDS are more concentrated and need more expertise for understanding.
Dialogues in TDS either focus on a single domain with more fine-grained topics, such as medical dialogues of different specialties,
or several pre-defined domains, such as restaurant, hotel, and transformation reservation.
Domain knowledge is significant for summarization, and it is divergent across sub-domains. For instance, expertise and medical knowledge are required in doctor-patient dialogues for generating accurate medical concepts~\cite{joshi2020dr} while specific knowledge bases for internal medicine and primary care are not the same.

\item The input dialogue for both ODS and TDS is made up of \textbf{a stream of utterance} as defined in Equation~\ref{eq:dialogue}. However, 
the \textbf{structure} of these two types of dialogues are different.
Open-domain dialogues often happen casually and freely while dialogues in TDS may have some inherent working procedures or writing formats. 
For example, the program manager in meetings usually masters the meeting progress~\cite{zhu2020end} implicitly with words such as ``okay, what about ...'', and communications by e-mails consist of semi-structured format including subjects, receivers, senders, and contents~\cite{zhang2021emailsum}. 

\item \textbf{Focuses of summaries} are distinct. Summaries for ODS in recent research are more like condensed narrative paraphrasing with different levels of granularity. An example is a synopsis from the Fandom wiki\footnote{\url{criticalrole.fandom.com}} maintained by fans for the Critical Role transcripts~\footnote{\url{github.com/RevanthRameshkumar/CRD3}}\cite{rameshkumar2020storytelling}, helping to quickly catch up with what is going on in the long and verbose dialogues. Differently, dialogues in TDS take place with strong intentions for solving problems. Summaries for such dialogues are expected to cover the user intents and corresponding solutions, such as medical summaries for clinical decision making~\cite{joshi2020dr} and customer service summaries for ticket booking~\cite{zhao2021todsum}. As a result, generating faithful content is extremely significant for TDS. 
\end{itemize}


\section{Overview of Approaches}
\label{sec:approach}

In abstractive text summarization, early researchers tried non-neural abstractive summarization methods~\cite{banko-etal-2000-headline}, which used statistical models to recognize important words and sentences and then concatenate them into a final summary with or without pre-defined templates. The most direct way is to select a set of keywords from input~\cite{nenkova2005impact}, such as log-likelihood ratio test~\cite{lin2000automated}, which identified the set of words that appear in the input more often than in a background corpus. Another way is to assign weights to all words in the input. Most popular such work relied on TF-IDF weights~\cite{berg2011jointly}. Word weights have also been estimated by supervised approaches with typical features, including word probability and location of occurrence~\cite{sipos2012large}. Some other traditional work directly focuses on predicting sentence importance, by either emphasizing select sentences that match the template of summaries or selecting the sentences in which keywords appeared near each other. Such sentences can better convey important information and be selected as a summary~\cite{celikyilmaz2010hybrid,litvak2010new}. Researchers also productively explored the relationship between word and sentence importance, and tried to estimate each in either supervised or unsupervised framework~\cite{liu2010supervised}. Since 2015, neural-based abstractive text summarization models~\cite{rush2015neural, nallapati2016abstractive, see2017get, liu2021keyword} began to be widely developed. These methods adapt recurrent neural network (RNN), convolutional neural network
(CNN) and Transformer architecture for sentence representation. Benefiting from the semantic representation learned from
neural networks and large training data, neural-based summarization methods outperform non-neural methods, especially in the aspect of fluency and semantic coherence.

\begin{figure}[t]
	\centering
	\subfigure[Sequential Modeling]{
		\begin{minipage}[t]{0.45\linewidth}
			\centering
			\includegraphics[scale=0.5]{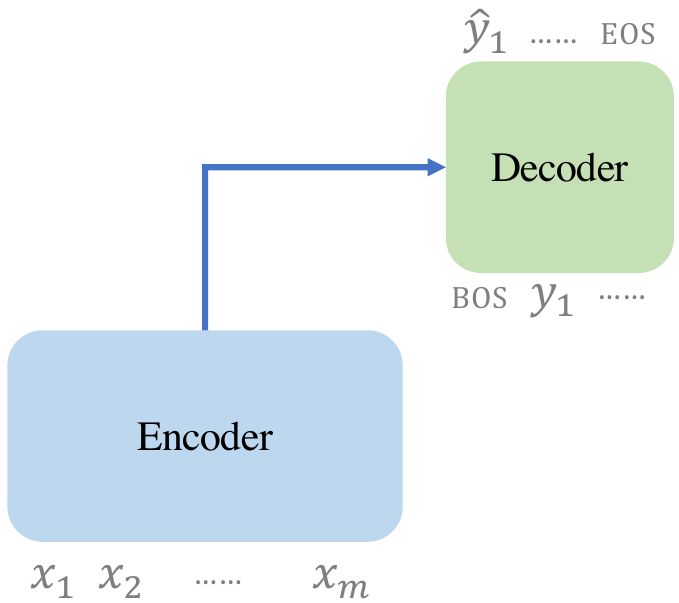}
		\end{minipage}
	}
	\subfigure[Hierarchical Modeling]{
		\begin{minipage}[t]{0.45\linewidth}
			\centering
			\includegraphics[scale=0.5]{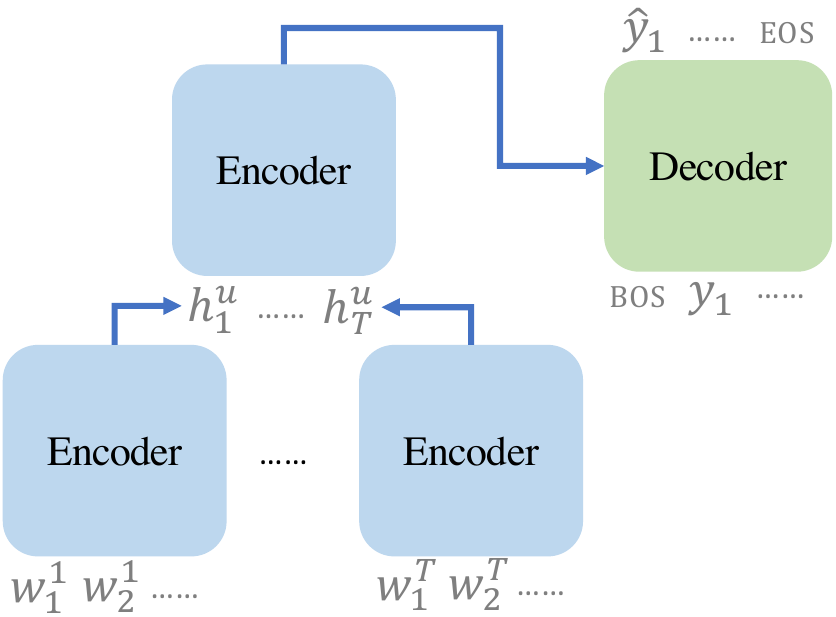}
		\end{minipage}
	}
	\caption{Two mainstream modeling designs for encoder-decoder summarization models.}
	\label{fig:encdec}
\end{figure}

The mainstream approaches in recent years hinge on the neural-based encoder-decoder architecture.  
In document/news summarization, 
document sentences can be concatenated into a single sequence of tokens $X=\{x_1, x_2, ..., x_m\}$ as the input to the encoder ${\rm Enc}(\cdot)$ which maps the tokens into 
contextualized hidden states $H = \{h_1, h_2, ..., h_m\}$. $m$ represents the number of input tokens. 
Besides such flat and sequential modeling, hierarchical modeling is another representative design as shown in Figure~\ref{fig:encdec}, which is usually favored by longer dialogues. Sentences are no more concatenated but instead modeled with hierarchical encoders. The lower layer encoder projects tokens within a sentence into hidden states. Then, the higher layer encoder takes these hidden states as sentence embeddings and projects them into global hidden representations.
The decoder ${\rm Dec}(\cdot)$ takes all of the hidden states $H$ and previously generated tokens 
as input, predicting the next token step by step in an auto-regressive way. The generation task is to minimize the negative log-likelihood $L$ with the teacher-forcing strategy as follows:
\begin{equation}
	\begin{aligned}
		&H= {\rm Enc}(x_1, x_2, ..., x_m) \\
		&P(y_p|y_{<p},H) = {\rm Softmax}(W_v{\rm Dec}(BOS, y_1, y_2, ..., y_{p-1}, H))\\
		&L = -\frac{1}{n}\sum_{p=1}^n P(y_p|y_{<p}, H) 
	\end{aligned}
\end{equation}
where $W_v$ is a trainable parameter matrix mapping hidden states into a vocabulary distribution. During inference, the predicted distribution over vocabulary at step $p$ is:
\begin{equation}
		P(\hat{y}_p|\hat{y}_{<p},H) = {\rm Softmax}(W_v{\rm Dec}(BOS, \hat{y}_1, \hat{y}_2, ..., \hat{y}_{p-1}, H))
\end{equation}
Tokens are sampled based on this distribution with generation strategies such as greedy and beam searches to produce the optimal generated summary. 
Greedy search selects the next token with the largest probability at each step and subsumes it into the current generation, while beam search expands each candidate generation with top-$k$ possible next tokens and preserves the $k$-best candidate generations at each step~\cite{rush2015neural}. The candidate with the highest probability is the final output.
The decoding process starts with the beginning of a sentence (BOS) token and terminates when the end of a sentence (EOS) token is generated. 
Basic neural architectures for encoders and decoders evolve from CNN~\cite{lecun1989handwritten,gehring2017convolutional} and RNN~\cite{rumelhart1985learning,nallapati2016abstractive} to Transformer~\cite{vaswani2017attention}. Nowadays, pre-trained models taking advantage of the Transformer encoder-decoder architecture with sequential modelings, such as BART and Pegasus, are the state-of-the-art abstractive text summarization techniques for document summarization.

These models also work for dialogue summarization.  
For sequential modeling, utterances prefixed with corresponding speakers are simply concatenated into the 
input sequence for a dialogue, i.e. 
\begin{equation}
	X = \{x_1, x_2, ..., x_m\} = [s_1, u_1, s_2, u_2, ..., s_T, u_T]
\end{equation}
where $[\cdot]$ represents concatenation operation.
However, such a simple operation largely ignores the flexible discourse structure and topic boundaries challenges in dialogue summarization.
For hierarchical modeling, utterances $\{u^t|_{t=1}^{T}\}$ are passed into encoders separately, which sets a significant barrier for word-level cross-utterance understanding.
Besides, models pretrained with normal text are not ideal for dialogue language understanding. 
To deal with these challenges as discussed in 
Section~\ref{sec:divergence}, a number of techniques have emerged. 
This survey mainly 
focuses on newly introduced techniques for adapting tested abstractive document summarization models to dialogues. 
More detailed explanations of
neural-based text summarization models and other methods please refer to
other surveys~\cite{shi2021neural,syed2021survey}.

\begin{figure*}
	\centering
	\includegraphics[scale=0.7]{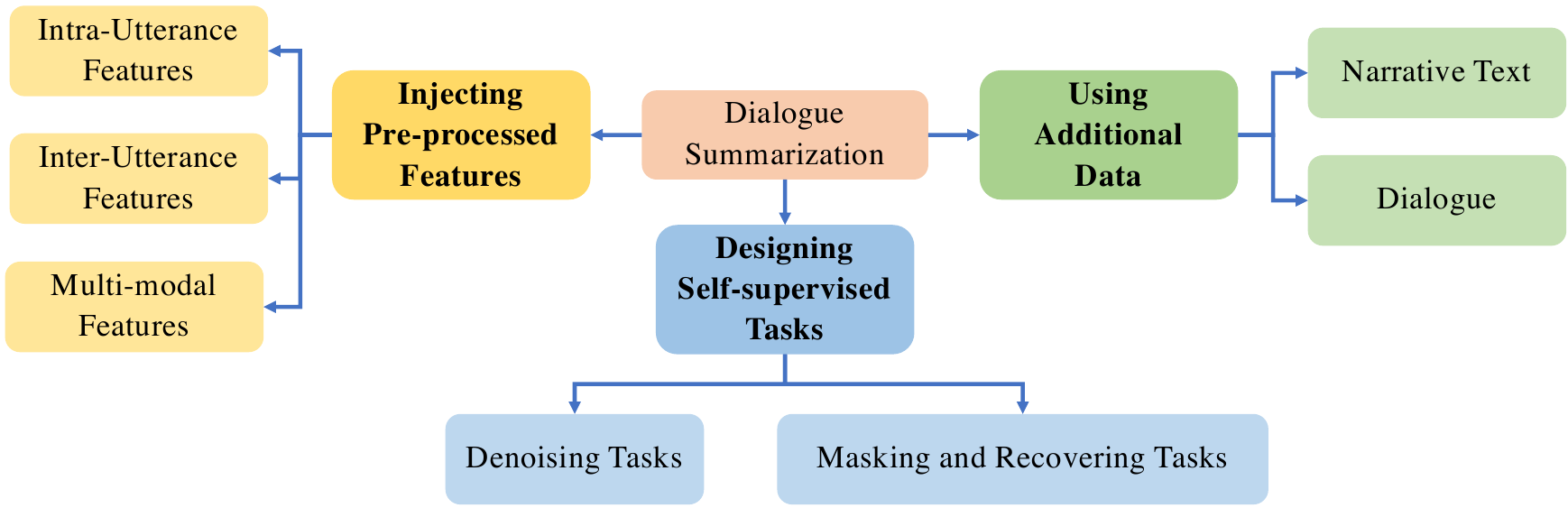}
	\caption{The taxonomy of dialogue summarization techniques. Methods are mainly categorized into three directions with more fine-grained sub-categories 
under each direction. Specific methods under each category are shown in 
white or gray boxes, but are not limited to these proposed options.}
	\label{fig:featuretaxonomy}
\end{figure*}

At a high level, recent researches tackle dialogue summarization in 
three directions: 
\begin{itemize}
\item \textbf{Injecting pre-processed features} which explicitly exploits additional features in dialogue context either by human annotators or external labeling tools as part of the input.
\item \textbf{Designing self-supervised tasks} which trains the model with auxiliary objectives besides the vanilla generation objective or individually for unsupervised summarization.
\item \textbf{Using additional data} which includes bringing training data from other related tasks or performing data augmentation based on existing training corpus.
\end{itemize}
A number of techniques have been proposed under each direction
which can be either adopted individually or combined for the targeted applications. An overall taxonomy is 
illustrated in Figure \ref{fig:featuretaxonomy}. The following three sections present more details about each direction, accompanied by highlights of pros and cons respectively.

\section{Injecting Pre-processed Features} 
\label{sec:feature}
To pursue better dialogue understanding and reasoning, different features either designed by experts back on linguistic knowledge or engineered with observations are proposed to simulate the human comprehension process.
Recognizing these features is not only independent dialogue analysis 
tasks but also critical enablers for downstream applications. 
A subset of these features has been proved helpful for dialogue summarization by extracting from $D$ explicitly and injecting it into the vanilla model.
We group different features into two sub-categories by their scopes:
\begin{itemize}
	\item \textbf{Intra-utterance features} are features within an utterance or for an individual utterance.
	\item \textbf{Inter-utterance features} are features connecting or distinguishing multiple utterances.
\end{itemize}

\subsection{Intra-Utterance Features}

We divide the intra-utterance features into three groups: word-level, 
phrase-level or utterance-level.

\subsubsection{Word-level}
Word-level intra-utterance features include TF-IDF weights, Part-of-speech (POS) tags, and named entity tags.

The \textbf{TF-IDF weight} is a well-known statistical feature for each word, signifying its importance in the whole corpus. 
Term-frequency, i.e., TF,  for each word is the number of repetitions in a dialogue or an utterance divided by the total number of words. Inverse-document-frequency abbreviated as IDF refers to the logarithm of the total number of dialogues or utterances divided by the number of them containing the word.
Each dialogue or utterance can be represented in a TF-IDF weight vector with a dimension of the vocabulary size, where each element is the product of TF and IDF. 
In early work, \citet{murray2005extractive} used such utterance vectors as features and input them to classifiers to find important utterances.
This feature is still prevalent in constructing better prompts for the summary generation with GPT-3~\cite{brown2020language}. Given a testing dialogue, \citet{prodan2021prompt} uses the cosine similarities between such dialogue vectors to find the most similar training dialogues for prompt construction.

\textbf{POS tags} and \textbf{named entity tags} are linguistic labels assigned for each word. POS tags represent grammatical properties, including nouns, verbs, adjectives, etc. Named entity tags belong to pre-defined categories such as person names, organizations, and locations. Both features are easily labeled by well-known NLP packages such as NLTK\footnote{\url{https://www.nltk.org/}} and Spacy\footnote{\url{https://spacy.io/}}. 
\citet{zhu2020end} trained two embedding matrices for both tags and concatenated them with word embeddings as part of the embedding layer for the model, i.e.
$x_i^t = [e_i^t;POS_i^t;ENT_i^t]$. 
$e_i^t$, $POS_i^t$, and $ENT_i^t$ are the word embedding, POS embedding, and named entity embedding for $x_i^t$, respectively. These features, which were also adopted by \citet{qi2021improving} in the same way, work for hierarchical models trained from scratch on this task and 
help with language understanding and entity recognition. 
However, the probing tests indicated that pre-trained 
language models have already captured both features well 
implicitly~\cite{tenney2018you,miaschi2020linguistic}, the two are no longer 
needed. The POS tags or dependency tags can also be assigned to summaries in the training set~\cite{OyaMCN14,singla2017spoken}, to generate summary templates for abstractive text summarization without neural models. 

\subsubsection{Phrase-level}
Phrase-level intra-utterance features have key phrases/words and negation scopes.
 
\textbf{Key phrases/words} emphasize salient 
n-grams in the original dialogue, which can help with the information scattering challenge and lead to more informative summaries. 
The definition of key phrases varies.
\citet{wu2021controllable} regarded the longest common sub-sequence (LCS) between each candidate phrase, extracted from $D$ first using a trained constituency parser, and $Y$ as key phrases. 
The LCSs are concatenated into a sketch, 
which is prefixed to $Y$ as a weakly 
supervised signal for the summary generation. 
Similarly, \citet{zou2021topic} proposed that words that appear both 
in $D$ and $Y$ are salient or informative topic words, i.e., another kind of keywords. 
They used an extension of the Neural Topic Model (NTM)~\cite{miao2017discovering} to learn the word-saliency correspondences. 
Then, input utterances are converted to topic representations by the saliency-aware NTM and further incorporated into Transformer Decoder layers for a better extractor-abstractor two-stage summarizer.
Differently, \citet{feng2021language} regarded unpredictable words by DialoGPT as keywords since they assumed that highly informative words could not be predicted. 
They appended all extracted keywords at the end of $X$ as inputs to the summarization model.

The \textbf{Negation scope} is also a set of consecutive words that 
reflect denied contents in utterances. \citet{chen2020multi} pointed out that negations are challenging for dialogues. With that in mind, \citet{khalifa2021bag} trained a Roberta model on CD-SCO dataset~\cite{morante2012sem} for negation scope prediction. This model labels the beginning and end positions of sentences' negation scopes in $D$ with designated special tokens. Unfortunately, inputting such labeled $D$ to the summarization model hurts the performance according to their experiment results. Negations are still of great importance in some task-oriented scenarios for generating accurate facts, such as realizing the patient's confirmation or negation of a symptom in a medical care conversation. \citet{joshi2020dr} proposed using an additional binary vector to label each $x_i$ based on a set of manually-curated negative unigrams, using this vector to modify the cross-attention distribution. Besides, they also extended the vocabulary with a special token `[NO]' and 
learned when to generate it by formulating the probability distribution over extended vocabulary, similarly to~\citet{see2017get}. The results showed reductions in coherency despite capturing negations, achieving competitive performance on doctor evaluations

\subsubsection{Utterance-level}
Speakers or roles, redundancies, user intents, and dialogue acts are utterance-level intra-utterance features. 
Domain knowledge is another kind of intra-utterance feature. It lies across phrase-level to utterance-level depending on specific circumstances. 

\textbf{Speaker} or \textbf{role} is a naturally provided ``label'' for each dialogue utterance. 
Since the general default input to models is the concatenation of all of 
the utterances into a sequence of tokens, each speaker or role token $s_t$ 
is encoded just like any other content token $w_i^t$~\cite{chen2020multi,feng2021language}. Thus, the speaker or role 
information is likely ignored or misunderstood, especially by language models pre-trained 
on common crawled texts. 
For a better understanding of speaker information, \citet{lei2021hierarchical} introduced Speaker-Aware Self-Attention made up of Self-Self Attention and Self-Others Attention to the vanilla Transformer layer, which only considered whether utterances were from the same speaker and avoided using the exact names. This structured feature is also 
adopted in~\cite{lei2021finer}.
In addition, the number of speakers is used as a feature for 
finding similar dialogues in the training set by \citet{prodan2021prompt}.
In TDS as mentioned in Section~\ref{sec:scenarios}, the number of roles is always fixed in a specific scenario, although the speakers are various among dialogue sessions.
\cite{yang2022tanet} modified the input with template ``\{\textit{participant}\} of role \{\textit{role}\} said: \{\textit{utterance}\}''.
Other previous work mainly focuses on modeling roles, reflecting functional information bias in utterances. 
The cheapest way is to represent each role with a dense vector $r_t$ which is either obtained by randomly initialized trainable vectors~\cite{zhu2020end,duan2019legal,qi2021improving,gan2021inspectional,asi2022end} or a small trainable neural network~\cite{song2020summarizing}. This vector is further concatenated, summed up, or fused by non-linear layers with input embeddings $e_i^t$ or utterance-level representations $h_t^u$ in summarization models. 
There are also works that capture such features by different sets of model parameters for different roles~\cite{zou2021topic,zhang2020unsupervised,yuan2019scaffolds}, which require a higher GPU memory footprint. More complicated methods that regard speakers or roles as graph nodes beyond the utterance-level will be introduced in Section~\ref{sec:graphs}.

Since dialogue utterances are mixed with backchanneling or repetitive confirmations~\cite{sacks1978simplest}, \textbf{redundancy} is also a significant 
utterance-level feature where each utterance is either preserved or removed. 
\citet{murray2005extractive} and \citet{zechner2002automatic} regarded 
utterances similar to the previous ones as redundant by calculating 
the cosine similarity between two sentence vectors 
computed using TF-IDF features. {Then, the remaining utterances can be regarded as a summary}.
Different from previous work calculating similarities between individual utterances,
\citet{feng2021language} brought the context into consideration which calculated similarities on the dialogue level.
Utterance representations $h_t^u$ are collected by inputting the whole dialogue into DialoGPT~\cite{zhang2020dialogpt} which is pretrained on 
dialogue corpus. 
Then, they assume that if adding an utterance $u_{t+1}$ to the 
previous history $\{u_1, ..., u_t\}$ doesn't result in a big difference 
between the context representation $h_t^u$ and $h_{t+1}^u$, 
$u_{t+1}$ will be regarded as a redundant utterance. 
Such features will be added as part of the dialogue input with special tokens. 
\citet{wu2021controllable} regarded non-factual utterances 
such as chit-chats and greetings as redundancies.
They used a sentence compression method with neural content selection to 
remove this less critical information as the first step for their summary 
sketch construction.

Another group of utterance-level features is matching each utterance with 
a label from a pre-defined multi-label set. \citet{wu2021controllable} defined 
a list of interrogative pronoun category to encode the \textbf{user intent}. 
Their definition is drawn upon the FIVE Ws principle and adapts to the 
dialogue scenario, including \textit{WHY}, \textit{WHAT}, \textit{WHERE}, 
\textit{WHEN}, \textit{CONFIRM} and \textit{ABSTAIN}. 
Each utterance is labeled by a few heuristics and these user intents are combined with the keywords and redundancies mentioned above as a sketch prefixed to the summary output.
This definition is different from the so-called user intent in task-oriented dialogue systems, while the latter can be used for TDS and will be discussed in domain ontologies in Section~\ref{sec:graphs}.

A more widely-accepted label set is \textbf{dialogue act}, which is defined as the functional unit used by speakers to change the context~\cite{bunt1994context} and has been used for different goals~\cite{kumar2018dialogue, oraby2017may}. The whole dialogue act taxonomy, including dialogue assess, inform, offer, etc., is tailored for different scenarios. For example, only 15 kinds of dialogue are labeled in the meeting summarization corpus AMI~\cite{carletta2005ami} 
while the total number of categories is 42~\cite{stolcke2000dialogue}. \citet{goo2018abstractive} explicitly modeled the relationships between dialogue acts 
and the summary by training both the dialogue act labeling task and 
abstractive summarization task jointly. 
\citet{di2020da} further added the dialogue act information as a contextualized 
weight to $h_t^u$. These labels are required from human annotators. 

\textbf{Commonsense knowledge} generated by widely-used generative commonsense model PARA-COMET~\cite{gabriel2021paragraph} is considered in~\cite{kim2022mind}. PARA-COMET takes dialogue history with the target utterance or a summary sentence as input and outputs short phrases for each of the 5 relation types, which are strongly correlated with speakers' intentions and the hidden knowledge, such as ``XINTENT'' and ``XREACT''. The generated knowledge is concatenated with each utterance as input and is used as the additional generation target in a dual-decoder setting.

In addition, \textbf{domain knowledge} plays an important role in TDS for 
dialogue understanding, even with pre-trained language models. 
\citet{koay2020meet} showed that the existence of 
terms affects summarization performance substantially. 
Such knowledge is considered as intra-utterance features in previous work.
\citet{joshi2020dr} leveraged a compendium of medical concepts for 
medical conversation summarization. They incorporated domain knowledge at 
the phrase level by simply encoding the presence of medical concepts, which 
are both in the source and the reference. The corresponding one-hot vectors 
affect the attention distribution by the weighted sum with contextualized 
hidden states $H$ for each word only during training, like the teacher forcing strategy.
\citet{gan2021inspectional} defined a number of domain aspects, and labeled text spans manually in $D$ and $S$. Auxiliary classification tasks of these aspects help generate more readable summaries covering important in-domain contents. 
Differently, work by~\citet{duan2019legal} incorporated their legal 
knowledge for each utterance. This is because their legal knowledge graph 
(LKG) depicts the legal judge requirements for different cases rather than a dictionary to look up, and each node represents a judicial factor 
requiring more semantic analysis beyond the word level. A series of graph 
knowledge mining approaches were adopted to seek relevant knowledge w.r.t. 
each utterance $u_t$, and the final legal knowledge embedding was added 
to the sentence embedding $h_t^u$ for further encoding.
 

\subsection{Inter-Utterance Features}\label{sec:interutt}

As dialogue utterances are highly dependent, information transitions among utterances are of great importance for dialogue context understanding. 
Multiple inter-utterance features show up for more efficient and 
effective dialogue summarization, which can be categorized into 
two sub-categories:
\begin{itemize}
	\item \textbf{Partitions} refer to extracting or segmenting the whole dialogue into relatively independent partitions. Information within each partition is more concentrated with fewer distractions for the summary generation. Meanwhile, these features reduce the requirements on GPU memory with shorter input lengths, which are especially preferred for long dialogue summarization.
	\item \textbf{Graphs} refer to extracting key information and relations from utterances to construct graphs, serving as a complement to the dialogue. These features are designed to help the summarization model understand the inherent dialogue structure.
\end{itemize}

\subsubsection{Partitions} 

There are two types of partitions.
One is to cut the dialogue into a sequence of $K$ consecutive segments $\{S_k|_{k=1}^K\}$ with or without overlaps, i.e., $|D|\leq|S_1| + ... + |S_K|$, where $|\cdot|$ counts the number of utterances. Representative features under this category are as follows.

\textbf{Topic transition} is an important feature for dialogues where speakers turn to focus on different topics. It has been studied as topic segmentation and classification~\cite{takanobu2018weakly}. Consecutive utterances that focus on the same topic constitute a topic segment, and the topic segment should meet three criteria\cite{arguello2006topic}, including being reproducible, not relying heavily on task-related knowledge,  and being grounded in discourse structure.
Some previous works annotate this feature when constructing datasets such as \citet{carletta2005ami} and \citet{janin2003icsi}. 
\citet{di2020da} took advantage of such labeled information during decoding.
Others collected such features by rules or algorithms.
\cite{asi2022end} adopted the text segmentation idea from~\cite{alemi2015text} and broke the long dialogue into semantically coherent segments by word embeddings.
\citet{liu2019topic} regarded different symptoms as different topics in medical dialogues and detected the boundaries by human-designed heuristics.
To alleviate human annotation burdens, unsupervised topic segmentation methods are adopted. \citet{chen2020multi} used the classic topic segmentation algorithm C99~\cite{choi2000advances} based on inter-utterance similarities, where utterance representations were encoded by Sentence-BERT~\cite{reimers2019sentence}. \citet{feng2021language} regarded sentences that are difficult to be generated based on the dialogue context to be the starting point of a new topic. Thus, sentences with the highest losses calculated based on DialoGPT~\cite{zhang2020dialogpt} are marked.
However, the window size and std coefficient for C99 algorithm~\cite{choi2000advances} in 
\citet{chen2020multi} and percentage of unpredictable utterances in \citet{feng2021language} are still hyper-parameters that need assigning by humans.
Among these works, some models use topic transitions as prior knowledge and input to summarisation models. They either add special tokens to dialogue inputs~\cite{chen2020multi,feng2021language}, add interval segment embeddings, such as $\{t_a, t_a, t_b, t_b, t_b, t_a,...\}$ for each utterance~\cite{qi2021improving}, or guide the model on learning segment-level topic representations $h_k^s$ based on utterance representations $h_t^u$~\cite{zheng2020abstractive}.
Others adjust their RNN-based models to predict topic segmentation first and do summarization based on the predicted topic segments~\cite{liu2019topic,li2019keep}, either with or without using additional supervised topic labels for computing the segmentation loss during training. 

Multi-view~\cite{chen2020multi} describes \textbf{conversation stages}~\cite{althoff2016large} from a conversation progression perspective. They assumed that each dialogue contained $4$ hidden stages, which were interpreted as ``openings$\rightarrow$intentions$\rightarrow$discussions$\rightarrow$conclusions'', and annotated with an HMM conversation model. These four stages. In their approach, both the preceding topic view and such stage view are labeled on dialogues with a separating token ``|'', encoded with two encoders sharing parameters and guided by the Transformer decoder in BART with additional multi-view attention layers. 

There also exists a simple \textbf{sliding-window} based approach that regards window-sized consecutive utterances as a snippet and collects snippets with different stride sizes.
On the one hand, it can be used to deal with long dialogues. Sub-summaries are generated for each snippet and merged to get the final summary.
On the other hand, pairs of (snippet, sub-summary) are augmented data for training better summarization models.
Most works regarded the window size and the stride size as two constants~\cite{koay2021sliding,liu2021topic,zhang2021leveraging,zhang2021summ},
while \citet{liu2021dynamic} adopted a dynamic stride size which predicts the stride size by generating the last covered utterance at the end of $Y'$.
\citet{koay2021sliding} generated abstractive summaries for each snippet by news summarization models as a coarse stage for finding the salient information.
Other work carefully matched the sentences in $Y$ with snippets to get 
better training pairs.
By calculating Rouge scores between reference sentences and snippets, 
the top-scored snippet is paired with the corresponding sentence~\cite{liu2021topic, zhang2021summ}.
Alternatively, multiple top-scored snippets can be merged as the corresponding input to the sentence~\cite{zhang2021leveraging} for the sub-summary generation. 
However, the gap between training and testing is that we don't know the oracle snippets since there is no reference summary during testing.
Therefore, each snippet was also considered to be paired with the whole summary~\cite{zhang2021leveraging,zhang2021summ}, but it leads to hallucination problems.
These constructed pairs can also be used with an auxiliary training objective~\cite{liu2021topic}, or as pseudo datasets for hierarchical summarization~\footnote{Hierarchical summarization means we do summarization, again and again, using the previously generated summaries as input to get more concise output. These models can either share parameters~\cite{li2021hierarchical} or not~\cite{zhang2021leveraging,zhang2021summ} in each summarization loop.}.

The other is to \textbf{cluster utterances} or \textbf{extract utterances} into a single part or multiple parts $\{P_l|_{l=1}^{K'}\}$. In this way, outlier utterances or unextracted utterances will be discarded, i.e., $|D|>|P_1| + ... + |P_K'|$. Then, the abstractive summarization model is trained between the partitions and the reference summary. The whole process can be regarded as variants under the extractor-abstractor framework for document summarization~\cite{chen2018fast,liu2021keyword}. 

\citet{zou2021unsupervised} proposed to select topic utterances according to 
centrality and diversity\footnote{Centrality reflects the center of utterance clusters in the representation space. Diversity emphasizes diverse topics among selected utterances.} in an unsupervised manner. Each utterance with its surrounding utterances in a window size forms a topic segment.
\citet{zhong2021qmsum} extracted relevant spans given the query with the Locator model which is initialized by Pointer Network~\cite{vinyals2015pointer} or a hierarchical ranking-based model. 
Cluster2Sent by~\citet{krishna2021generating} extracted important utterances, clustered related utterances together and generated one summary sentence per cluster, resulting in semi-structured summaries suitable for clinical conversations. 
\citet{banerjee2015generating} and \citet{shang2018unsupervised} followed a similar procedure, including (segmentation, extraction,  summarization) and (clustering, summarization) respectively.
The oracle spans are required to be labeled for supervised training of extractors or classifiers for most approaches, except that \citet{shang2018unsupervised} used K-means for utterance clustering in an unsupervised manner.
Generally, the partitions are concatenated as the input to abstractive summarization models~\cite{zhong2021qmsum}, or the generated summary of each segment is concatenated or ranked to form the final $\hat{Y}$~\cite{zou2021unsupervised,banerjee2015generating}.


\subsubsection{Graphs} 
\label{sec:graphs}

The intuition for constructing graphs is attributed to the divergent structure between dialogues and documents mentioned in Section~\ref{sec:divergence}. To capture the semantics among complicated and flexible utterances, a number of works constructed different types of graphs based on different linguistic theories or observations and demonstrated improvements in dialogue summarization tasks empirically. We group these graphs into three categories according to the type of 
nodes in the graph, i.e., being either a word, a phrase or an utterance. 

\textit{Word-level graphs} focus on finding the central words buried in the whole dialogue. Some works~\cite{OyaMCN14,banerjee2015generating,shang2018unsupervised,park2022unsupervised} parsed utterances together with or without summary templates using the Standford or NLTK packages. Words in the same form and the same POS tag or synonyms, according to WordNet~\cite{MehdadCTN13} are regarded as a single node. Either the natural flow of text, parsed dependency relations or relations in WordNet are adopted to connect nodes, resulting in a directed \textbf{word graph}. It is used for unsupervised sentence compression by selecting paths covering nodes with high in-degree and out-degree without language models.

Complex interactions within dialogues always make it hard for humans and models to associate speakers with correct events. At the same time, different surface forms for the same event and frequent coreferences increase the difficulty for the model to generate faithful summaries. The purpose for \textit{phrase-level graphs} is mainly to emphasize relations between important phrases.
\citet{liu2021coreference} and \citet{liu2021controllable} transferred document coreference resolution models~\cite{joshi2020spanbert,lee2018higher} to dialogues,  applied data post-processing with human-designed rules and finally constructed \textbf{coreference graphs} for dialogues. The nodes are mainly person names and pronouns, and the edges connect the nodes belonging to the same mentioned cluster.
Based on the coreference results, \citet{chen2021structure} took advantage of information extraction system~\cite{angeli2015leveraging} and constructed an \textbf{action graph} with "WHO-DOING-WHAT" triples. ``WHO'' and ``WHAT'' constitute nodes, and the direction of edges representing ``DOING''  is from ``WHO'' to ``WHAT''.
\citet{zhao2021give} manually defined an undirected \textbf{semantic slot graph} based on NER and POS Tagging focusing on entities, verbs, and adjectives in texts, i.e., slot values. Edges in this graph represent the existence of dependency between slot values collected by a dependency parser tool.
More strictly defined ``domain-intent-slot-value'' tuples based on structured \textbf{domain ontologies} are marked in advance~\cite{yuan2019scaffolds,zhao2021todsum}. It is different from domain to domain, such as ``food, area'' slots 
for ``restaurant'' and ``leaveAt, arriveBy'' slot for ``taxi'' labeled in 
the MultiMOZ dataset~\cite{eric2019multiwoz}.
Ontologies in the medical domains containing clinical guidelines in 
``subject-predicate-object'' triples were introduced in~\citet{molennar2020healthcare}'s work. Triples are extracted from $D$ and matched with the ontology to construct a patient medical graph for report generation.
Moreover, external commonsense \textbf{knowledge graphs}, 
such as ConceptNet~\cite{speer2012representing}, have also been adopted to find the relations among speaker nodes, utterance nodes and knowledge nodes~\cite{feng2021incorporating}. The graph is undirected with ``speaker-by'' edges connecting speaker nodes and utterance nodes and ``know-by'' edges connecting utterance nodes and knowledge nodes.

\textit{Utterance-level graphs} considering the relationship between utterances have been explored mainly in four ways. 
One is \textbf{discourse graph} mainly based on the SDRT theory~\cite{asher2003logics} which models the relationship between elementary discourse units (EDUs) with 16 types of relations for dialogues. Both \citet{chen2021structure} and \citet{feng2020dialogue} adopted this theory and regarded each utterance as an EDU. They labeled the dialogue based on a discourse parsing model~\cite{shi2019deep} trained on a human-labeled multi-party dialogue dataset~\cite{asher2016discourse}. 
The former work used a directed discourse graph with utterances as nodes and discourse relations as edges.
Differently, the latter one transformed the directed discourse graph with the Levi graph transformation where both EDUs and relations are nodes in the graph with two types of edges, including default and reverse.  Self edges and global edges were also introduced to aggregate information in different levels of granularity.
\citet{ganesh2019restructuring} designed a set of discourse labels themselves and trained a simple CRF-based model for discourse labeling. Unfortunately, they haven't released the details about the discourse labels so far.
\textbf{Dependency graph} can be regarded as a simplification of discourse graph since it only focuses on the ``reply-to'' relation among utterances.
The tree structure of a conversation is a kind of it and is adopted in~\cite{yang2022tanet} by modifying the self-attention into thread-aware attention which considers the distance between two utterances, and also proposing a thread prediction task to predict the historical utterances in the same thread for some sampled utterances.
Another one is \textbf{argument graph}~\cite{stede2016parallel} for identifying argumentative units, including claims and premises and constructing a structured representation. \citet{fabbri2021convosumm} did argument extraction with pretrained models~\cite{chakrabarty2019ampersand} and connected all of the arguments into a tree structure for each conversation by relationship type classification~\cite{mirko2020towards}. Such a graph not only helps to reason between arguments but also eliminates unnecessary content in dialogues. Similarly, \textbf{entailment graph}~\cite{MehdadCTN13} is also used to identify important contents by entailment relations between utterances.
The fourth is \textbf{topic graph}. Usually, we regard the topic structure in dialogues as a linear structure as discussed above, but it can be hierarchical with subtopics~\cite{carletta2005ami, janin2003icsi} or other non-linear structures since the same topic may be discussed back and forth~\cite{kim2019dynamic}.
\citet{lei2021finer} used ConceptNet to find the related words in dialogue. These words indicate the connections among utterances under the same topic, capturing more flexible topic structures.

The graphs above have been used for dialogue summarization in three ways. One is to convert the original dialogue into a narrative format similar to documents by linearizing graphs and inputting to the basic summarization models~\cite{fabbri2021convosumm,ganesh2019restructuring} introduced in Section~\ref{sec:approach}.
Second is to bring graph neural layers, such as Graph Attention Network~\cite{velivckovic2018graph} and Graph Convolutional Networks~\cite{kipf2016semi}, for capturing the graph information. Such graph neural layer can be solely used as the encoder~\cite{feng2021incorporating}. It can also cooperate with the Transformer-based encoder-decoder models, either based on the hidden states from the encoder or injected as a part of the Transformer layer in encoder~\cite{liu2021coreference} or decoder~\cite{chen2021structure}.
The rest modify attention heads in the Transformer architecture with the constructed graphs from a model pruning perspective. \citet{liu2021coreference} replace the attention heads that represent the most coreference information with their coreference graph, while \citet{liu2023picking} replace the underused heads with similar coreference graphs.

\subsection{Multi-modal Features}

Humans live and communicate in a multi-modal world. As a result, multi-modal dialogue summarization is naturally expected. Even for virtual dialogues from TV shows or movies, character actions and environments in videos are important sources for humans to generate meaningful summaries. However, due to the difficulties of collecting multi-modal data in real life and the limited multi-modal datasets, this area remains to be researched. Only 
\textbf{prosodic features} gained attention in early speech-related works, which contribute to automatic speech recognition (ASR). For example, 
\citet{murray2005extractive} collect the mean and standard deviation of the fundamental frequency, energy and duration features based on speech. The features are collected at the word level and then averaged over the utterance.
With the marvelous ASR models, most works later only focused on transcripts and ignored such multi-modal features.
Besides, \textbf{visual focus of 
attention} {(VFOA)} feature from the meeting summarization scenarios 
has been introduced to highlight the importance of utterances~\cite{li2019keep}. 
It represents the interactions among speakers reflected by the focusing target that each participant looks at in every timestamp. They assumed that the longer a speaker was paid attention to by others, his or her utterance would be more important. Such orientation feature was converted into a vector by their proposed VFOA detector framework and further concatenated to the utterance representations. 

\subsection{Summary and Opinions}
\begin{figure}
	\centering
	\includegraphics[scale=0.7]{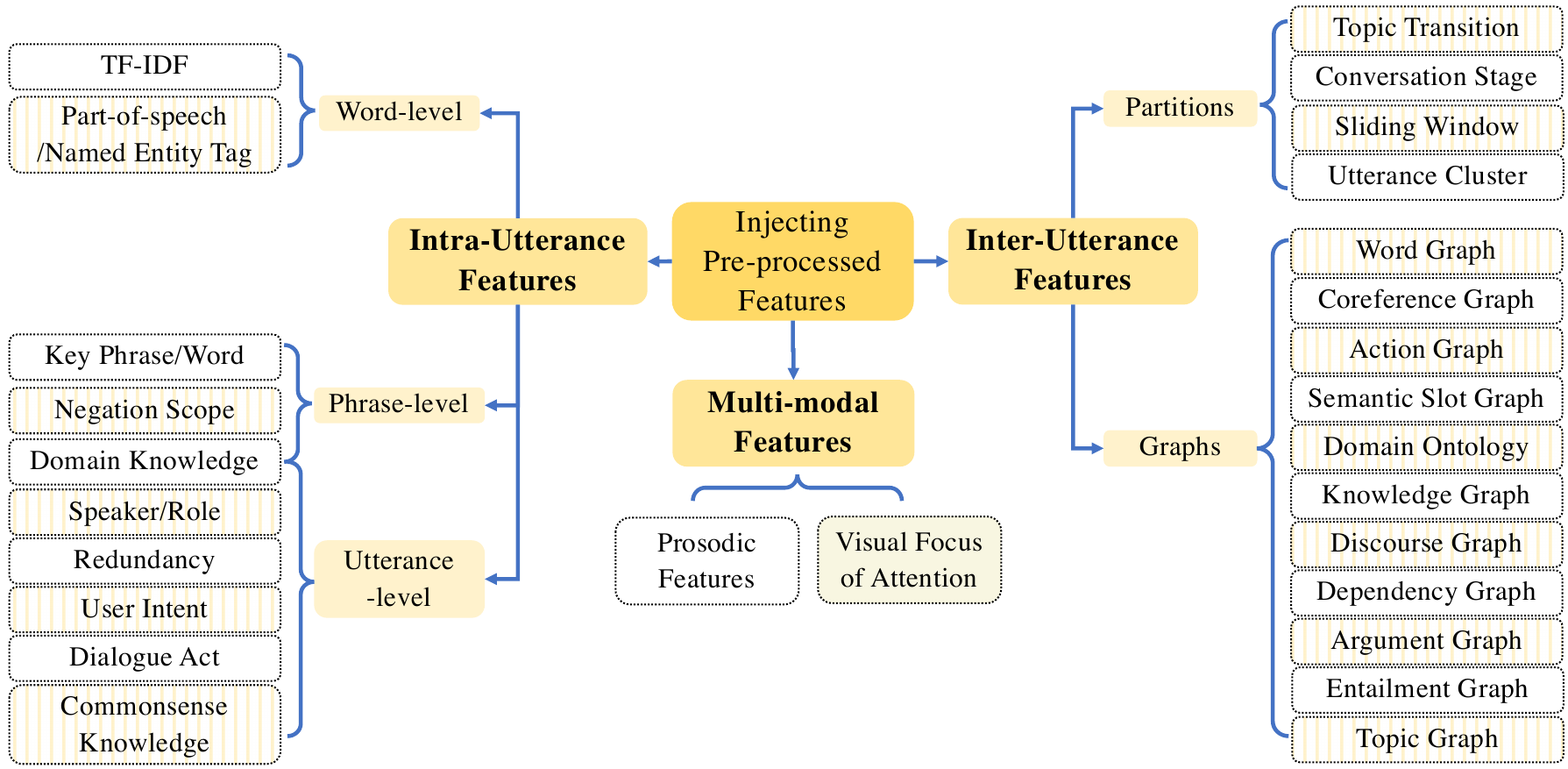}
	\caption{A summary of all features.}
	\label{fig:app-feature}
\end{figure}
The features mentioned above are summarized in 
Figure~\ref{fig:app-feature}. They are mainly {injected into vanilla models} 
in three ways: 

\begin{itemize}
\item \textbf{Manipulating the input and output} by adding annotations or data reformulation. The former one adds additional tokens to the dialogue or summary to highlight the corresponding features, such as topic transition marks in the dialogue~\cite{chen2020multi} and key phrase prefixes of the summary~\cite{wu2021controllable}. This is suitable for features linearly buried in the texts. For the hierarchical or more complicated structures, researchers tend to reformulate the dialogue into different segments, especially for long dialogues~\cite{zhong2021qmsum,banerjee2015generating,shang2018unsupervised} or reordering utterances considering graph features. For example,  \citet{fabbri2021convosumm} linearized the argument graph following a depth-first approach to train a graph-to-text summarization model based on pre-trained sequence-to-sequence language models. Besides, \citet{zhao2021todsum} linearized the final dialogue states, i.e., slot-related labels, as a replacement of $D$ with a bi-encoder model.


\item \textbf{Modifying the model architecture or hidden states} for learning inductive bias on known features. Embedding layers are always modified for word-level or phrase-level features indicating the binary or multi-class classification properties, including the POS embeddings in~\cite{zhu2020end} and medical concept embedding in~\cite{joshi2020dr}. Modifications on self-attentions and cross-attentions are used to merge multiple features and are also preferable to graph features. For instance, \citet{chen2020multi} modified the cross-attention layer for balancing and fusing hidden states of two kinds of labeled input from double encoders. \citet{lei2021hierarchical} changed the self-attention layer in the encoder with two speaker-aware attentions to highlight the information flow within the same speaker or among speakers. Different graph neural layers~\cite{feng2021incorporating,liu2021coreference,chen2021structure} are also introduced for capturing graph features.

\item \textbf{Adding additional training targets} means that features are regarded as a supervision output during training under multi-task learning and are ignored during inference. For example, \citet{goo2018abstractive}, \citet{li2019keep}, \citet{kim2022mind} used an additional decoder for dialogue act labeling, topic segmenting and commonsense knowledge generation, respectively.
\citet{yuan2019scaffolds} incorporated domain features by formulating domain classification as a multi-label binary classification problem for the whole $D$. All of them are using utterance-level features to help learn better encoder representations, which will lead to a high-quality summary in the end.

\end{itemize}

The advantages and disadvantages of injecting pre-processed features are as follows:
\begin{itemize}
    \item[\Checkmark] Injecting pre-processed features as the mainstream research direction for dialogue summarization significantly improves the results compared with the basic summarization model. Features including negation scope, speaker/role, coreference graph, action graph and semantic slot graph pay more attention to generating consistent summaries, while most of the other features help to select valuable information for summarization.
    \item[\Checkmark] Such explicitly incorporated features are more interpretable to humans and can be manipulated for more controllable summaries. Different features can be selected and combined to promote the model performance in specific application scenarios.
        \item[\Checkmark] Features collected by labelers on other dialogue understanding tasks capture the essence of these tasks and also establish connections with various aspects of dialogue analysis.  Therefore, leveraging such features is a good way to alleviate the human labeling burden.
	\item[\XSolidBrush] Features are not transferable in different scenarios and some features are not compatible with each other, thus feature engineering is shown to be important.
	\item[\XSolidBrush] Labelers trained with other datasets are always out-of-domain compared to the targeting dialogue summarization scenario. Hyper-parameters introduced in labeling algorithms with these labelers need try and error for the domain transfer. 
	\item[\XSolidBrush] Error propagation exists in these dialogue summarization approaches. Incorrect features hinder the understanding of dialogues and lead to poor summaries. 
\end{itemize}


{

\section{Designing Self-supervised Tasks}\label{sec:designselftasks}

To alleviate human labor and avoid error propagation, self-supervised 
tasks emerged, which leverage the dialogue-summary pairs without additional 
labels.  We divide such tasks used in recent works into three sub-categories:
\begin{itemize}
	\item \textbf{Denoising tasks} which are designed for eliminating noises 
in the input or penalizing negatives during training.
	\item \textbf{Masking and recovering tasks} which means that parts of the input are masked and the masked tokens are required to be predicted.
	\item \textbf{Dialogue tasks} which refer to response selection and generation tasks for better dialogue understanding.
\end{itemize}
Specific works are as follows.

\subsection{Denoising Tasks}

Denoising tasks focus on adding
noises to the dialogue input or output and aims at generating concise summaries by filtering out the noisy information, which results in more robust dialogue 
summarization models.
\citet{zou2021unsupervised} used the original dialogue as output and trained a denoising auto-encoder which is capable of doing content compression for unsupervised dialogue summarization.
Noising operations, which include fragment insertion, utterance replacement, and
content retention, are applied together on each sample.
For a utterance $u_t$ in $D$, \textbf{fragment insertion} means that randomly sampled 
word spans from $u_t$ is inserted to $u_t$ for lengthening the original 
sequence.
\textbf{Utterance replacement} is that $u_t$ is replaced by another utterance $u_{t'}$ in $D$ and \textbf{content retention} means that $u_t$ is unchanged.} 
\citet{chen2021simple} augmented dialogue data by swapping, deletion, insertion 
and substitution on utterance level and used the corresponding summary as 
the output, resulting in more various dialogue inputs for training the 
dialogue summarization model. 
\textbf{Swapping} and \textbf{deletion} aim to perturb discourse relations by randomly swapping two utterances in $D$ or randomly deleting some utterances.
\textbf{Insertion} includes inserting repeated utterances that are chosen from $D$ randomly and inserting utterances with specific dialogue acts such as self-talk or hedge from a pre-extracted set, aiming for mimicking interruptions in natural dialogues and generating more challenging inputs.
\textbf{Substitution} replaces the chosen utterances in $D$ by utterances generated with a variant of text infilling task adopted in the BART pre-training process.
Different from \citet{zou2021unsupervised}, only one operation is adopted to noise $D$ at a time, and these operations pay more attention to dialogue characteristics, such as the structure and context information.

This kind of task can be extended to learn beyond the denoising ability when 
combined with contrastive learning or classification tasks on positive and negative data. 
Contrastive learning trains the model to maximize 
the distance between positive data and negative data for learning more informative semantic representations, which extends the classification's ability on generation tasks.
\citet{liu2021topic} 
proposed {coherence detection} and {sub-summary generation} 
for implicitly modeling the topic change and handling information scattering 
problems. They cut the dialogue into snippets by sliding windows and 
separated the long summary into sentences as a first step.
\textbf{Coherence detection} is to train the encoder to distinguish 
a snippet with shuffled utterances from the original ordered one.
\textbf{Designated sub-summary generation} is to train the model to generate more related 
summaries by constructing negative samples with unpaired dialogue 
snippets and sub-summaries, where the positive pair is obtained 
by finding the snippet with the highest Rouge-2 recall for each sub-summary.
The loss is calculated according to generation losses.
\citet{tang2021confit} also designated summaries where negative summaries are constructed for different error types, such as swapping the nouns for wrong reference and object errors, swapping verbs for circumstance errors and tense and modality errors, etc. Positive summaries are collected by back translation technology. The distance of decoder representations measures the contrastive loss.
They also considered the \textbf{token identification} task to determine whether two tokens belong to the same speaker according. Encoder representations of tokens are used for classification.
\citet{zhao2021give} made improvements by \textbf{perturbing hidden representations} of 
the target summary for alleviating the exposure bias following~\citet{lee2020contrastive}, which has been proven to be useful for conditional generation tasks.

\subsection{Masking and Recovering Tasks} 

Masking and recovering tasks 
are commonly used in pre-training for better language modeling by recovering the original dialogue and 
bears some resemblance to the noising operations. The main difference is that these tasks try to recover the original text given the corrupted one.
 It can be divided into 
work-level and sentence-level by the granularity of masked contents.
Word-level masks for \textbf{pronouns}~\cite{khalifa2021bag}, \textbf{entities}~\cite{liu2022entity,khalifa2021bag}, 
\textbf{high-content tokens}~\cite{khalifa2021bag}, 
\textbf{roles}~\cite{qi2021improving} and \textbf{speakers}~\cite{zhong2021dialoglm} are 
considered in previous work, for a better understanding of the complicated speaker 
characteristics and capturing salient information. Words masked 
in \citet{khalifa2021bag}'s work was determined by POS tagger, 
named entity recognition or simple TF-IDF features. Although the lexical 
features and statistical features have been captured by pre-trained models 
for different words as mentioned in Section~\ref{sec:feature}, predicting 
the specific content words under these features reversely given the dialogue context is still challenging and helpful to dialogue context modeling especially with models pre-trained on general text.
Utterance-level masking objective inspired by \textbf{Gap Sentence Prediction}~\cite{zhang2020pegasus} is adopted by~\citet{qi2021improving}. Differently, key sentence selection from dialogues is done by a graph-based sorting algorithm TextRank and Maximum Margin Relevance. 
\citet{zhong2021dialoglm} introduced three new utterance-level tasks, 
including turn splitting, turn merging, and turn permutation. 
\textbf{Turn splitting} is cutting a long utterance into multiple turns and 
adding ``[MASK]'' in front of each turn except the first one with 
the speaker. \textbf{Turn merging} is randomly merging consecutive turns 
into one turn and neglecting the speakers except the first one. 
And \textbf{turn permutation} means that utterances are randomly shuffled.
All of these tasks are trained to recover the original dialogue by predicting the masked words or changed utterances.

\subsection{Dialogue Tasks} 

There are also papers incorporating
well-known {dialogue tasks} into dialogue summarization. General \textbf{response 
selection} and \textbf{generation} models can be trained with unlabelled dialogues by 
simply regarding a selected utterance $u_t$ as the output and 
the utterances before it $u_{<t}$ as the input. Negative candidates for 
the selection task are the utterances randomly sampled from the whole corpus.
\citet{fuzw20} assumed that a superior summary is a representative of the 
original dialogue. So, either inputting $D$ or $Y$ is expected to achieve 
similar results on other auxiliary tasks. This way, the next utterance generation 
and classification tasks {acted like evaluators, to give guidance on better summary generation}.
\citet{feigenblat-etal-2021-tweetsumm-dialog} trained response selection models for identifying salient utterances.
The intuition is that the removal of a salient utterance in dialogue context 
will lead to a dramatic drop in response selection, 
and these salient sentences are the same for summarization. 
This way, they regard the drop in probability as a saliency score to 
rank the utterances and adopt the top 4 utterances as the
extractive summary, which can also be further used to enhance abstractive 
results by appending it at the end of the dialogue as the input.

\subsection{Summary and Opinions}

\begin{figure}
	\centering
	\includegraphics[scale=0.7]{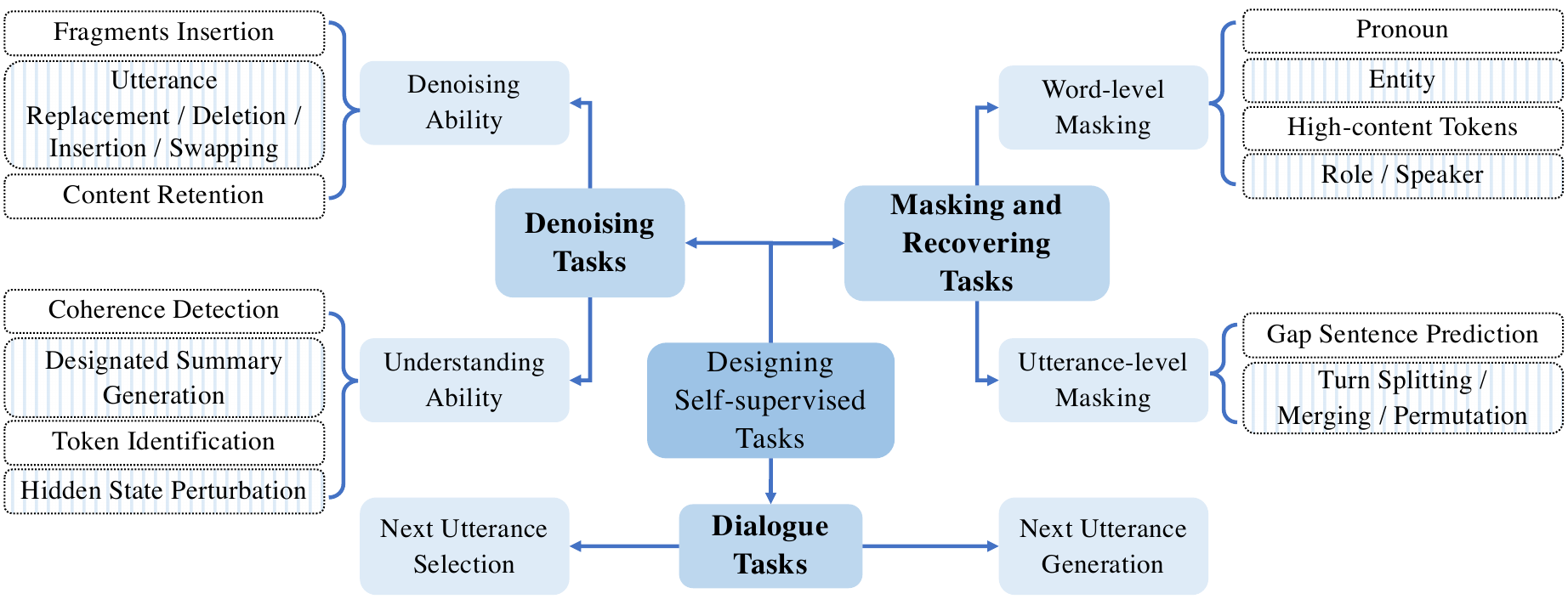}
	\caption{A summary of self-supervised tasks.}
	\label{fig:app-task}
\end{figure}

The tasks mentioned above are in Figure~\ref{fig:app-task}. Most of these self-supervised tasks are adopted in two ways:
\begin{itemize}
	\item \textbf{Cooperating with the vanilla generation task under 
different training paradigms.}	
{Multi-task learning refers that the losses from self-supervised tasks are weighed summed with the vanilla generation for updating}~\cite{zhao2021give,fuzw20}, or updated sequentially in a batch~\cite{liu2021topic}. 
Pre-training with auxiliary tasks and then fine-tuning on dialogue summarization with the vanilla generation task is also widely accepted in~\cite{khalifa2021bag,qi2021improving}. 
The former is usually selected when the auxiliary training tasks are close to the summarization target. 
The latter one is chosen for learning more general representations, which is also more flexible to use additional data in Section~\ref{sec:useadddata}. 
	\item \textbf{Training an isolated model for different purposes.} 
The model is used as the summarization model directly~\cite{zou2021unsupervised, feigenblat-etal-2021-tweetsumm-dialog}, or as a trained labeler providing information for dialogue summarization~\cite{feigenblat-etal-2021-tweetsumm-dialog} with less artificial facts compared with \citet{feng2021language}. 

\end{itemize}

The advantages and disadvantages of designing self-supervised tasks are as follows:
\begin{itemize}
	\item[\Checkmark] Most self-supervised tasks take advantage of 
self-supervision to train the model. They don't need to go through
the expensive and time-consuming annotation process for collecting 
high-quality labels, and avoid the domain transfer problems 
of transferring labelers trained on the labeled domain to the target 
summarization domain.
	\item[\Checkmark] Useful representations are learned with these tasks by the summarization model directly or as an initial state for the summarization model, avoiding the error propagation caused by wrong labels. Although labeling tools such as POS tagger and TextRank are adopted, these predicted labels are not used as the training target or explicitly injected into the summarization model. They are just incorporated to find more effective self-supervisions.
	\item[\Checkmark] It's a good way to make full use of dialogue-summary 
pairs without additional labels, or even utilize pure dialogues without 
summaries. The latter is especially beneficial to unsupervised 
dialogue summarization.
	\item[\XSolidBrush] Although designing 
self-supervised tasks reduces the data pre-processing complexity, 
it increases the training time and computing costs for training the model 
with additional training targets 
on corresponding variations of the data.
	\item[\XSolidBrush] Different self-training tasks are not always 
compatible and controllable. It is challenging to design suitable tasks for dialogue summarization and 
find the best combination of tasks in different scenarios.
\end{itemize}

\section{Using Additional Data}\label{sec:useadddata}

Since dialogue summarization data is limited, researchers 
adopt data augmentation or borrow datasets from other tasks. 
We divide the data into two categories: 
Narrative Text and Dialogues.

\subsection{Narrative Text}

A number of {narrative text corpora} are utilized to do language modeling and learn commonsense knowledge which is shared across tasks.
Since most of today's summarization models are based on pre-trained encoder-decoder models, such as BART~\cite{lewis2020bart}, PEGASUS~\cite{zhang2020pegasus}, and T5~\cite{raffel2020exploring},  \textbf{common crawled text corpora} can be regarded as the backbone corpora of dialogue summarization. It generally includes Wikipedia, BookCorpus~\cite{zhu2015aligning} and 
C4~\cite{raffel2020exploring}. 
These pre-trained models, such as GPT-3~\cite{brown2020language}, can be directly used for dialogue summarization with prefix-tuning approaches~\cite{prodan2021prompt}.
\citet{li2021learn} transformed such data by dividing the sequence 
into two spans, selecting span pairs with higher overlaps by Rouge scores for training their model with better copying behaviors.
Overlapped text generation task is proposed which uses the first span to generate the second span. It further boosts their proposed model with the correlation copy mechanism on both document and dialogue summarization tasks, which copies words from $D$ on better occasions during the summary generation.

Document summarization is the most similar task to dialogue summarization. As a result, \textbf{document summarization data} are a natural choice for learning the summarization ability.
\citet{zhang2021exploratory} show that BART pre-trained with 
CNN/DM~\cite{hermann2015teaching}~\footnote{\url{https://huggingface.co/facebook/ bart-large-cnn}} enhances the dialogue summarization in the meeting and drama scenarios.
CNN/DM, Gigaword~\cite{rush2015neural}, and 
NewsRoom~\cite{grusky2018newsroom} were all adopted to train a model from scratch by~\citet{zou2021low}.
For taking advantage of models trained document summarization data and doing zero-shot on dialogues, \citet{ganesh2019restructuring} narrowed down the format gap between documents and dialogues by restructuring dialogue to document format with complicated heuristic rules, such as discourse labels mentioned in Section~\ref{sec:graphs}
Differently, \citet{zhu2020end} shuffled sentences from multiple 
documents to get a simulated dialogue for pre-training, including 
CNN/DM, XSum~\cite{narayan2018don} and NYT~\cite{evan2008nyt}.
Similarly, \citet{park2022leveraging} simulated dialogues based on each document with three transformation functions: arranging text into dialogue format by adding ``Speaker 1:'', shuffling sentence order and omitting the most extractive sentence for enhancing the abstractiveness of constructed samples.

\textbf{Commonsense knowledge data} are also welcomed since it is a
basis for language understanding.
\citet{khalifa2021bag} considered three reasoning tasks, including ROC stories 
dataset~\cite{mostafazadeh2016corpus} for short story ending prediction, 
CommonGen~\cite{lin2020commongen} for generative commonsense reasoning, 
and ConceptNet for commonsense knowledge base construction. 
These three tasks, together with dialogue summarization, are jointly trained with multi-task learning and show a performance boost.

Besides, MSCOCO~\cite{lin2014microsoft} as a \textbf{short text corpus} is used in~\citet{zou2021low} for training the decoder with narrative text generation ability.

\subsection{Dialogue}

For collecting or constructing more \textbf{dialogue summarization data} 
without the need for human annotations, data augmentation approaches are 
proposed. \citet{liu2021controllable} and \citet{khalifa2021bag} augmented 
by replacing person names in both the dialogue and the reference summary 
at the same time. These augmented data are definitely well-paired and 
are preferred to mixing with the original data during fine-tuning. 
\citet{jia-etal-2022-post} simply paired the whole dialogue with each summary sentence and further trained the model with a prefix-guided generation task before fine-tuning, where the first several tokens of the target sentence are provided for guiding the model on generation and learning to rephrase from dialogue to narrative text to some extent.
\citet{asi2022end} shows that it is possible to take advantage of giant language models such as GPT-3~\cite{brown2020language} to collect pseudo summaries by inputting dialogues with pre-defined question hints.
\citet{liu2022data} collects augmented training pairs with a small seed dataset by following steps: aligning summary spans with utterances, replacing utterances by reconstruction of the masked dialogue,
and filling up the masked summary given the augmented dialogue. 
All these three steps are done with trained models on other tasks having sufficient data.
\cite{fang2022spoken} augmented and refined the original training pairs with an utterance rewriter model~\cite{liu2020incomplete} and a coreference resolution model~\cite{joshi2020spanbert}, leading to understandable dialogues.
Besides, using relatively large-scaled crawled dialogue summarization data as 
a pre-training dataset, such as MediaSum~\cite{zhu2021mediasum}, 
for other low-resource dialogue summarization scenarios was considered in \citet{zhu2021mediasum}'s work. 
For crawled data without summaries, pseudo summaries are constructed by selecting leading comments from the long forum threads on the Reddit~\cite{yang2022tanet}.

Other \textbf{dialogue data} without paired summary are also valuable.
\citet{feng2020dialogue} took questions as outputs and a number of utterances after each question as inputs, regarding question generation as the pre-training objective to help identify important contents in downstream summarization. 
\citet{khalifa2021bag} adopted word-level masks mentioned 
before on PersonaChat~\cite{zhang2018personalizing} and Reddit comments 
for fine-tuning. 
\citet{qi2021improving} pre-trained with dialogues from MediaSum and TV4Dialogue besides document summarization datasets used in \cite{zhu2020end}. They also stitch dialogues randomly to simulate topic transitions. 
\citet{zhong2021dialoglm} proposed a generative pre-training framework for long dialogue understanding and comprehension. Different from BART, Pegasus or T5 pretrained on general common crawled text, DialogLM in this paper is pretrained on dialogues from MediaSum dataset and OpenSubtitles Corpus~\cite{lison2016opensubtitles2016}. It corrupts a window of dialogue utterances with dialogue-inspired noises, similar to the noising operations mentioned in Section~\ref{sec:designselftasks}. 
The original window-sized utterances are the recovering target based on the remaining dialogue. Such a window-based recovering task is proposed to be more suitable for dialogues considering its scattered information and highly content-dependent utterances. 
Besides, \citet{amanda-etal-2022-shift} took advantage of a self-annotated corpus based on SAMSum~\cite{gliwa2019samsum} which converts each utterance individually to a third-person rephrasing.
They showed benefits on the same dataset under the zero-shot setting by pre-training with the perspective shift corpus.

Furthermore, \citet{zou2021low} broke the training for dialogue summarization 
model into three parts, namely encoder, context encoder and decoder, 
to train the dialogue modeling, summary language modeling and 
abstractive summarization respectively. 
Dialogue corpus, short text corpus, and summarization corpus were all 
used in this work, helping to bridge the gap between out-of-domain 
pre-training and in-domain fine-tuning, especially for low-resource settings.
	
\subsection{Summary and Opinions}

\begin{figure}
	\centering
	\includegraphics[scale=0.7]{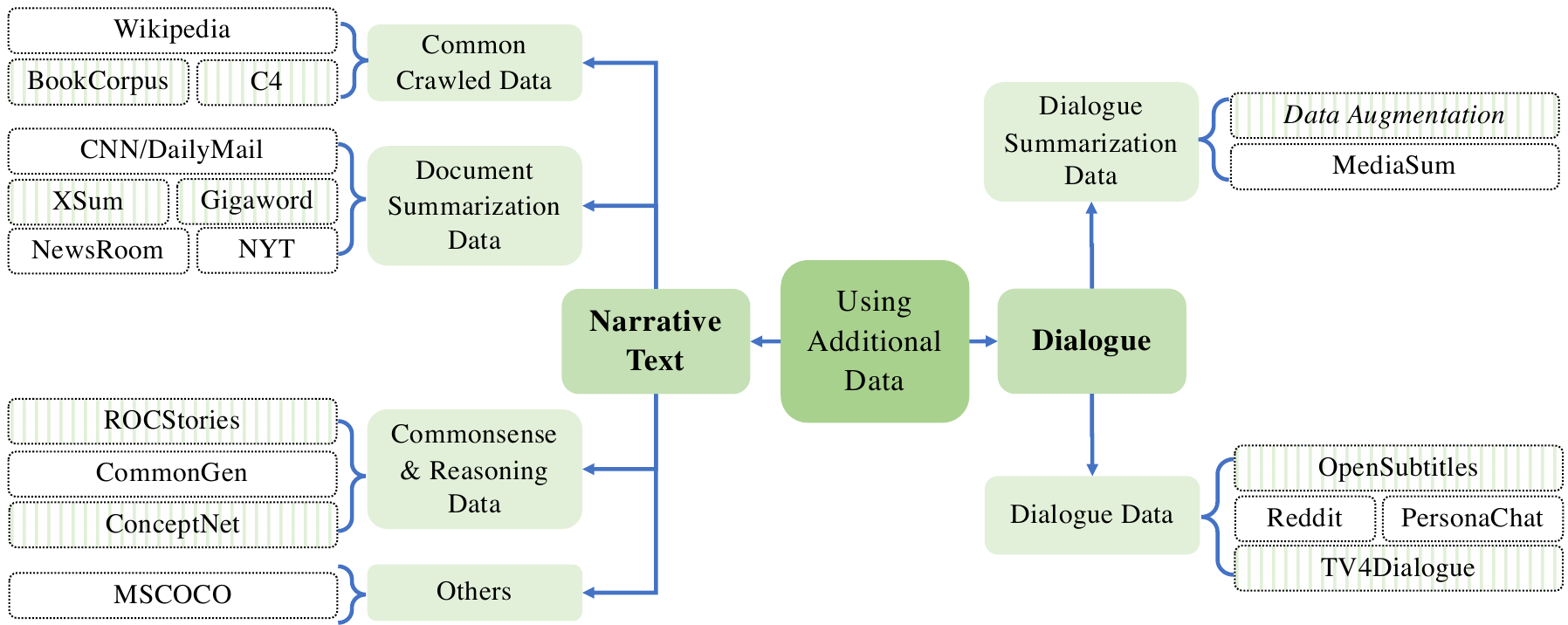}
	\caption{A summary of additional data.}
	\label{fig:app-data}
\end{figure}

Additional data in previous work are summarized in Figure~\ref{fig:app-data}. These data are always used in the following ways:
\begin{itemize}
	\item \textbf{Pre-training with corresponding training objectives}. 
Common crawled text data, document summarization data and dialogue data are 
mostly used in this way~\cite{zou2021low}, where the language styles or data 
formats are quite different from dialogue-summary pairs. It hopes to 
provide a better initialization state of the model for dialogue summarization. 
On the other hand, It is also a good way of coarse-to-fine-grained training, 
where pre-training is done with the noisy data by data augmentation or from other 
domains and fine-tuning with the oracle dialogue summarization training data~\cite{feng2020dialogue,zhu2021mediasum}.

\item \textbf{Mixing with dialogue summarization training data} and 
training for dialogue summarization directly. 
Data here are usually more similar to dialogue-summary pairs obtained by data 
augmentation~\cite{liu2021controllable,khalifa2021bag} or with intensive 
commonsense~\cite{khalifa2021bag,liu2022data}. 
\end{itemize}

The advantages and disadvantages of using additional data are as follows:
\begin{itemize}
	\item[\Checkmark] The language understanding ability among different 
corpora is the same intrinsically. As a result, additional data helps 
dialogue summarization, especially in low-resource settings, 
which further alleviates the burden of summary annotation by humans.
	\item[\Checkmark] The intensive knowledge in specially designed 
corpora helps strengthen the dialogue summarization model. 
	\item[\Checkmark] The additional unlabeled data can be trained 
with self-supervised tasks mentioned in Section~\ref{sec:designselftasks} for 
better performance.
	\item[\XSolidBrush] Training with additional data makes significant improvements while requiring 
more time and computational resources, demonstrating the sample inefficiency of current dialogue summarization models.
	\item[\XSolidBrush] Training with more data is not always effective~\cite{zhang2021exploratory,nair2021data}, especially when the divergence between 
the additional corpus and original dialogue summarization corpus is huge.  Elaborate data augmentation approaches avoid this problem when training data is not too scarce.
\end{itemize}

\section{Evaluations}
\label{sec:evaluation}

In this section, we present a comprehensive description of existing dialogue summarization datasets 
under different scenarios and introduce several widely-accepted evaluation 
metrics for this task.

\subsection{Datasets}
\label{sec:dataset}

A great number of dialogue summarization datasets have been proposed from different resources. We categorize them according to the scenarios in Section \ref{sec:scenarios}. 

\subsubsection{Open-domain Dialogue Summarization}

Open-domain dialogue summarization datasets under daily chat, drama conversation and debate\&comment are as follows and summarized in Table~\ref{tab:open}.

\textit{Daily Chat Datasets}: \textbf{SAMSum}~\cite{gliwa2019samsum} and \textbf{DialogSum}~\cite{chen2021dialsumm} are two large-scale real-life labeled datasets. Each dialogue in SAMSum is written by one person to simulate a real-life 
messenger conversations and the single reference summary is annotated by 
language experts. DialogSum, on the other hand, contains dialogues from 
the existing dialogue dataset, including DailyDialog~\cite{li2017dailydialog}, 
DREAM~\cite{sun2019dream} and MuTual~\cite{cui2020mutual}, and other English-speaking practice websites. These spoken dialogues have a more formal style than those in SAMSum, and each is accompanied by three reference summaries in the test set.  
Besides, AIHub Dialogue Summarization Dataset (\textbf{HubDial})~\footnote{https://aihub.or.kr/} also contains dialogues covering a range of daily topics.

\textit{Drama Conversation Datasets}: \textbf{CRD3}~\cite{rameshkumar2020storytelling} is collected from a live-stream role-playing game called Dungeons and Dragons, which is more amenable to extractive approaches with low abstractiveness.
 \textbf{MediaSum}~\cite{zhu2021mediasum} includes interview transcripts from 
NPR and CNN and their reviews or topic descriptions are regarded as the 
corresponding summaries. The large size of this automatically crawled 
dataset makes it particularly suitable for pre-training. 
Other two datasets are collected from a variety of movies and TV series, 
including \textbf{SubTitles}~\cite{malykh2020sumtitles} and 
\textbf{SummScreen}~\cite{chen2021summscreen}. Dialogues are corresponding 
transcripts, and summaries are aligned synopses or recaps 
written by humans.
 
 \textit{Debate\&Comment Datasets}: \textbf{ADSC}~\cite{misra2015using} 
is a test-only dataset extracted from the Internet Argument 
Corpus~\cite{walker2012your}. It contains 45 two-party dialogues about gay 
marriages, each  associated with 5 reference summaries. 
\textbf{FORUM}~\cite{tarnpradab2017toward} contains human-annotated forum threads collected from tripadvisor.com and ubuntuforums.org.
Three out of four sub-datasets in \textbf{ConvoSumm}~\cite{fabbri2021convosumm} 
are similar discussions, including news article comments (\textbf{NYT}), 
discussion forums and debate (\textbf{Reddit}) and community question answers 
(\textbf{Stack}) from different sources. Each sample has a human-written reference.
\textbf{CQASUMM}~\cite{chowdhury2019cqasumm} is another community question 
answering dataset but without back and forward discussions among speakers. The summary here aims to summarize multiple answers, which is closer to a multi-document summarization setting.

\begin{table}[th]
	\centering
	\small
		\begin{tabular}{|l|c|c|c|c|c|p{4cm}|c|}
			\hline
			\textbf{\makecell[c]{Name}} & \textbf{\makecell{$\#$Samples \\ train/val/test}} & \textbf{$\#$Spk} & \textbf{Lang.} & \textbf{DW} & \textbf{SW} & \textbf{\makecell[c]{Download Link}} & \textbf{AVL} \\
			\hline
			\multicolumn{6}{|l|}{\bf \em{Daily Chat}} \\
			\hline
			SAMSum\cite{gliwa2019samsum} & 14.7k/0.8k/0.8k
			& $\geq$2 & English & 94 & 25 & \tabincell{l}{https://huggingface.co/datasets\\/samsum}& Y \\
			\hline
			DialogSum\cite{chen2021dialsumm} & 12.5k/0.5k/0.5k
			& 2& English & 131 & 22 &\tabincell{l}{https://github.com/cylnlp/\\DialogSum} & Y\\
			
			\hline
			HubDial & 350k & $\geq$2 & Korean & - & - &\tabincell{l}{https://aihub.or.kr/}  & C \\
			
			\hline
			\multicolumn{6}{|l|}{\bf \em{Drama Conversation}} \\
			\hline
			CRD3\cite{rameshkumar2020storytelling} &	26.2k/3.5k/4.5k 
			& $\geq$2 & English & 31,803 & 2,062 & \tabincell{l}{https://github.com/\\RevanthRameshkumar/CRD3}& Y \\
			\hline
			MediaSum\cite{zhu2021mediasum} &
			463.6k/10k/10k 
			& $\geq$2 & English & 1,554 & 14 & \tabincell{l}{https://github.com/\\zcgzcgzcg1/MediaSum/}& Y \\
			\hline
			\makecell[l]{SumTitles\cite{malykh2020sumtitles}\\(Subtitiles/Scripts/Gold)} & \makecell[c]{132k\\21k\\290}
			& $\geq$2 & English & \makecell[c]{6,406\\423\\395} & \makecell[c]{85\\55\\51} & \tabincell{l}{https://github.com/huawei-\\noah/noah-research/tree/\\master/SumTitles}& Y \\
			\hline
			\makecell[l]{SummScreen\cite{chen2021summscreen}\\(FD/TMS)} &\makecell[c]{3,673/338/337\\18,915/1,795/1,793} 
			& $\geq$2 & English & \makecell[c]{7,605\\6,421} & \makecell[c]{114\\381} & \tabincell{l}{https://github.com/mingdachen\\/SummScreen}& Y \\
			\hline
			\multicolumn{6}{|l|}{\bf \em{Debate \& Comment}} \\
			\hline
			ADSC\cite{misra2015using} & 45 & 2 & English & 672 & 151 &\tabincell{l}{https://nlds.soe.ucsc.edu/\\summarycorpus}& Y \\
			\hline
			CQASUMM\cite{chowdhury2019cqasumm} & 100k
			& $\geq$2 & English& 782 & 100 &\tabincell{l}{https://bitbucket.org/tanya1410\\9/cqasumm/src/master/} & Y\\
			
			\hline
			FORUM~\cite{tarnpradab2017toward} & 689 & $\geq$2 & English & 825 & 191 &  \tabincell{l}{http://tinyurl.com/jcqgcu8} & Y \\
			
			\hline
			\makecell[l]{ConvoSumm\cite{fabbri2021convosumm}\\(NYT/Reddit/Stack)} &  \makecell[c]{-/0.25k/0.25k\\-/0.25k/0.25k\\-/0.25k/0.25k}
			& $\geq$2 &  \tabincell{l}{English}& \makecell[c]{1,624\\641\\1,207} & \makecell[c]{79\\65\\73} & \tabincell{l}{https://github.com/\\Yale-LILY/ConvoSumm} &Y \\
			
			\hline
			
		\end{tabular}
		\caption{Open-domain dialogue summarization datasets. ``Lang.''  and ``Spk'' stands for ``Language'' and ``Speakers''. ``DW'' and ``SW'' represents the average number of words in the dialogues and summaries respectively. ``AVL'' refers to the public availability of the
dataset ($Y$ is available, $N$ is not available, and $C$ is conditional). HubDial is only available for Koreans.}
		\label{tab:open}		
\end{table}

\subsubsection{Task-oriented Dialogue Summarization}

Datasets here are rooted in specific domains, including
customer service, law, medical care and official issue. We list them in Table~\ref{tab:task}. 

\textit{Customer Service Datasets}: Zou et al.\shortcite{zou2021topic,zou2021unsupervised} propose two similar datasets with summaries from the agent perspective.
\citet{lin2021csds} provides a more fine-grained dataset \textbf{CSDS} containing a user summary, an agent summary, and an overall summary based on JDDC dataset~\cite{chen2020jddc}. 
Summaries from \textbf{Didi dataset}~\cite{liu2019automatic} are also written from agents' points of view, in which dialogues are about transportation issues instead of pre-sale and after-sale topics in the former one.
More complicated multi-domain scenarios are covered in \textbf{TWEETSUMM}~\cite{feigenblat-etal-2021-tweetsumm-dialog}, \textbf{MultiWOZ*}~\cite{yuan2019scaffolds} and \textbf{TODSum}~\cite{zhao2021todsum}. Dialogues from TWEETSUMM spread over a wide range of domains, including gaming, airlines, retail, and so on. 
MultiWOZ* and TODSum transform and annotate summaries based on the original MultiWOZ dataset~\cite{eric2019multiwoz}.
There are also two earlier datasets called \textbf{DECODA} and \textbf{LUNA}~\cite{favre2015call} containing call centre conversations with synopses summarizing the problem of the caller and how it is solved.  

\begin{table}[t]
	\centering
	\small		
		\begin{tabular}{|l|c|c|c|c|c|p{3.7cm}|c|}
			\hline
			\textbf{\makecell[c]{Name}} &\textbf{ \makecell{$\#$Samples \\ train/val/test}}& \textbf{$\#$Spk} & \textbf{Lang.} & \textbf{DW} & \textbf{SW} & \textbf{\makecell[c]{Download Link}} & \textbf{AVL} \\
			\hline
			\multicolumn{6}{|l|}{\bf \em{Customer Service}} \\
			
			\hline
			\citet{zou2021topic} & 17.0k/0.9k/0.9k
			& 2 & Chinese & 1,334 & 55 &\tabincell{l}{https://github.com/RowitZou\\/topic-dialog-summ}& Y \\
			
			\hline
			CSDS\cite{lin2021csds} & 9.1k/0.8k/0.8k
			& 2& Chinese & 401 & 83 &\tabincell{l}{https://github.com/xiaolin\\Andy/CSDS} & Y\\
			
			\hline
			{\citet{zou2021unsupervised}} & -/0.5k/0.5k
			& 2 &  \tabincell{l}{Chinese}& 95 & 37 & \tabincell{l}{https://github.com/RowitZou\\/RankAE} &Y \\
			
			\hline
			{Didi\cite{liu2019automatic}} &296.3k/2.9k/29.6k 
			& 2 & Chinese & - & - &	\tabincell{l}{-}& N \\
			
			\hline
			{TWEETSUMM\cite{feigenblat-etal-2021-tweetsumm-dialog}} & 0.9k/0.1k/0.1k 
			& 2 & English & 245 & 36 & \tabincell{l}{https://github.com/guyfe\\/Tweetsumm}& Y \\

			\hline
			MultiWOZ*\cite{yuan2019scaffolds} & 8.3k/1k/1k & 2 & English & 181 & 92 & \tabincell{l}{https://github.com/voidforall\\/DialSummar}& Y\\
			
			\hline
			{TODSum\cite{zhao2021todsum}} & 9.9k & 2 & English & 187 & 45 &\tabincell{l}{-}& N \\
			
			\hline
			DECODA\cite{favre2015call} & -/50/100 & 2 & \makecell[c]{French/\\English}
			& \makecell[c]{42,130\\41,639} & \makecell[c]{23\\27} & \tabincell{l}{https://pageperso.lis-lab.fr/\~benoit\\.favre/cccs/} & C\\
			
			\hline
			LUNA\cite{favre2015call} & -/-/100 & 2 & \makecell[c]{Italian/\\English}
			& \makecell[c]{34,913\\32,502} & \makecell[c]{17\\15}  &\tabincell{l}{https://pageperso.lis-lab.fr/\~benoit\\.favre/cccs/}  & C\\
			
			\hline
			\multicolumn{6}{|l|}{\bf \em{Law}} \\
			
			\hline
			{Justice\cite{fuzw20}} & 30k
			& 2 & Chinese & 605 & 160 & \tabincell{l}{-}& N \\
			
			\hline
			{PLD\cite{duan2019legal}} & 5.5k& $\geq$2 & English  & - & - &\tabincell{l}{https://github.com/zhouxinhit\\/Legal\_Dialogue \_Summarization} & C \\
			
			\hline
			{LCSPIRT-DM\cite{xi2020global}} &  30.8/3.8k/3.8k
			& 2 &  Chinese& 684 & 75 & \tabincell{l}{http://eie.usts.edu.cn/prj/\\NLPoSUST/LcsPIRT.htm} & C \\
		
			\hline
			\multicolumn{6}{|l|}{\bf \em{Medical Care}} \\
		
			\hline
			{\citet{joshi2020dr}} & 1.4k/0.16k/0.17k
			& 2 & English & - & - &\tabincell{l}{-}& N \\
			
			\hline
			{\citet{song2020summarizing}} & 36k/-/9k 
			& 2& Chinese  & 312 & 23/113 &\tabincell{l}{https://github.com/cuhksz-nlp\\/HET-MC} & Y\\
			
			\hline
			{\citet{liu2019topic}} & 100k/1k/0.49k
			& 2 &  \tabincell{l}{English}& - & - & \tabincell{l}{-} &N \\
			
			\hline
			{\citet{zhang2021leveraging}} & 0.9k/0.2k/0.2k 
			& 2 & English & - & - & \tabincell{l}{-}& N \\
			
			\hline
			\multicolumn{6}{|l|}{\bf \em{Official Issue (Meeting \& Emails)}} \\
			
			\hline
			{AMI\cite{carletta2005ami}} &137 
			& $>$2 & English & 4,757 & 322 & \tabincell{l}{https://groups.inf.ed.ac.uk/ami}& Y \\
			
			\hline
			{ICSI\cite{janin2003icsi}} & 59 
			& $>$2 & English & 10,189 & 534 &\tabincell{l}{https://groups.inf.ed.ac.uk/ami\\/icsi}& Y \\
			
			\hline
			{QMSum\cite{zhong2021qmsum}} & 1.3k/2.7k/2.7k
			& $>$2 & English & 9070 & 70 &\tabincell{l}{https://github.com/Yale-LILY\\/QMSum}& Y \\
			
				\hline
			{Kyutech\cite{yamamura2016kyutech,nakayama2021corpus}} &  9 
			& $>$2 & Japanese & - & - &\tabincell{l}{http://www.pluto.ai.kyutech.\\ac.jp/~shimada/resources.html}& Y \\
			
			\hline
			{BC3\cite{ulrich2008publicly}} & 30
			& $>$2 & English & 550 & 134 &\tabincell{l}{https://www.cs.ubc.ca/cs-\\research/lci/research-groups\\/natural-language-processing\\/bc3.html} & Y \\
			
			\hline
			{\citet{loza2014email}} & 107
			& $>$2 & English & - & - &\tabincell{l}{-} & N\\
			
			\hline
			{EmailSum\cite{zhang2021emailsum}} & 1.8k/0.25k/0.5k
			& $\geq$2 & English& 233 & 27/69 &\tabincell{l}{https://github.com/ZhangShiyue\\/EmailSum} & C \\
			
			\hline
			\makecell[l]{ConvoSumm\cite{fabbri2021convosumm}\\(Email)} &  -/0.25k/0.25k%
			& $\geq$2 &  \tabincell{l}{English} &917 & 74 & \tabincell{l}{https://github.com/Yale-LILY\\/ConvoSumm} &Y \\
			
			\hline
		
		\end{tabular}	
		\caption{Task-oriented dialogue summarization datasets. The original text data is not accessible for PLD due to privacy issues. DECODA, LUNA and LCSPIRT-DM can only be obtained through an application. EmailSum is not free.}
		\label{tab:task}
\end{table}

\textit{Law Datasets}: \textbf{Justice}~\cite{fuzw20} includes 
debates between a plaintiff and a defendant on some controversies 
which take place in the courtroom. The final factual statement by the 
judge is regarded as the summary.
A similar scenario is included in \textbf{PLD}~\cite{duan2019legal}, which is more 
difficult to summarize due to the unknown number of participants. There is also another version 
of PLD by~\citet{gan2021inspectional} with fewer labeled cases than the 
original PLD.
\citet{xi2020global} proposed a long text summarization dataset \textbf{LCSPIRT-DM} based 
on police inquiry records full of questions and answers.

\textit{Medical Care Datasets}:
Both \citet{joshi2020dr} and \citet{song2020summarizing} proposed medical summarization corpora by crawling data from online health platforms and annotating coherent summaries by doctors. \citet{song2020summarizing} also proposed one-sentence summaries of medical problems uttered by patients, whereas \citet{liu2019topic} used simulated data with summary notes in a very structured format.
 \citet{zhang2021leveraging} used unreleased dialogues with coherent summaries of the history of the present illness. 

\textit{Official Issue Datasets}: \textbf{AMI}~\cite{carletta2005ami} and \textbf{ICSI}~\cite{janin2003icsi} are meeting transcripts concerning 
computer science-related issues in working background and research background, respectively. Both datasets are rich in human labels, including extractive summary, abstractive summary, topic segmentation, and so on. They are also included in \textbf{QMSum}~\cite{zhong2021qmsum} and are further labeled for query-based meeting summarization. \textbf{Kyutech}~\cite{yamamura2016kyutech} is a similar dataset in Japanese containing multi-party conversations, where the participants pretend to be managers of a virtual shopping mall in a virtual city and do some decision-making tasks. Their later work~\cite{nakayama2021corpus} annotated more fine-grained summaries for each topic instead of the whole conversation in ~\cite{yamamura2016kyutech}.
In addition, official communications are also prevalent in e-mails. 
\citet{ulrich2008publicly} propose the first email summarization dataset \textbf{BC3} with only 30 threads and \citet{loza2014email} release 107 email threads. Both of them contain extractive as well as abstractive summaries.
EmailSum~\cite{zhang2021emailsum} has both a human-written short summary and a long summary for each e-mail thread. 
Besides, Email threads (\textbf{Email}) in ConvoSumm~\cite{fabbri2021convosumm} have only one abstractive summary for each dialogue.

\subsubsection{Summary}
We make the following observations and conclusions.
\begin{itemize}
	\item The size of dialogue summarization datasets is much smaller than document summarization datasets. Most dialogue summarization datasets have no more than $30K$ samples, while representative document summarization datasets, such as CNNDM and XSum, have more than $200K$ samples. Datasets for drama conversations are relatively larger and can be potential pre-training data for other scenarios.
	\item The number of interlocutors in different dialogue summarization scenarios is different. Most ODS dialogues have more than $2$ speakers while 
most dialogues in TDS have only 2 speakers except in official meetings or 
e-mails.
	\item TDS dialogues tend to be more private. Thus, half of the 
TDS datasets are not publicly available, especially for Law and 
Medical Care scenarios. 
	\item Datasets with more than 2,048 dialogue words, which is the upper bound of the input length of most pre-trained language models, are suitable for research on long dialogue summarization. They contain both open-domain datasets and task-oriented datasets. 
\end{itemize}


\subsection{Evaluation Metrics}
\label{sec:evalmetric}
In existing works, \textit{Automatic evaluation metrics} commonly used for summarization such as \textbf{Rouge}~\cite{lin2004rouge}, \textbf{MoverScore}~\cite{zhao2019moverscore}, \textbf{BERTScore}~\cite{zhang2019bertscore} and \textbf{BARTScore}~\cite{yuan2021bartscore} are also used for dialogue summarization by comparing the generations with references. However, these widely-accepted metrics' performance may deviate from human~\cite{chen2021dialsumm,hanna2021fine}, especially in the aspect of consistency. Therefore, more focussed evaluation metrics and human evaluations emphasizing \textit{information coverage} and \textit{factual consistency} are considered as follows.

Instead of comparing only with the whole reference summary, most researches for TDS only consider key words/phrases
while ignoring other common words for measuring the \textbf{information coverage}.  In other words, evaluation for TDS emphasizes the coverage of key information which are generally domain-specific terms and can be easily recognized.
For example, {medical concept coverage}~\cite{joshi2020dr,zhang2021leveraging} 
and {critical information completeness}~\cite{yuan2019scaffolds} both
extract essential phrases based on domain dictionaries by 
rules or publicly available tools. 
\citet{zhao2021give} uses slot-filling model~\cite{chen2019bert} to recognize slot values for {factual completeness}.
Then, the accuracy or F1 scores are 
calculated by comparing extracted phrases or concepts from $Y$ and $Y'$.


ODS pays less attention to information coverage due to the higher subjectivity on salient information selection. Instead, measuring the \textbf{factual consistency} of generations gains increasing attention. Unlike the above metrics which compare generations with the reference summary, 
most evaluation metrics here compare generations with the source dialogue and can be classified into reference-free evaluation metrics~\cite{shao2017efficient,durmus2020feqa,egan2022play,liu2022reference}.
A QA-based model~\cite{wang2020asking} is borrowed by \citet{zhao2021give}.
It follows the idea that factually consistent summaries and documents generate the same answers to a question.
NLI-based methods~\cite{maynez2020faithfulness} that require the content in the summary to be fully inferred from the dialogue were adopted by~\citet{liu2022data}.
\citet{liu2021controllable} automatically evaluate {inconsistency} issues 
of person names by using noised reference summaries as negative samples and training a BERT-based binary classifier.
\citet{asi2022end} used the FactCC metric from~\citet{kryscinski2020evaluating} where the model was trained only with source documents with a series of rule-based transformations.
Information correctness of the generated summary is also important for TDS. For instance, negation correctness as a specific consistency type is considered by ~\citet{joshi2020dr} with 
publicly available tools Negex~\cite{harkema2009context} for recognizing 
negated concepts.

Meanwhile, \textit{human evaluations} are required to complement the above metrics.
Besides ranking or scoring the generated summary with an overall quality score~\cite{chen2020multi}, 
more specific aspects are usually provided to annotators. Representative ones include:
\textbf{readability/fluency}~\cite{yuan2019scaffolds,zhao2021give} requiring a summary to be grammatically correct and well structured,
\textbf{informativeness}~\cite{feng2020dialogue,lei2021finer,feigenblat-etal-2021-tweetsumm-dialog,feng2021language} measuring how well the summary includes salient information,
\textbf{conciseness/non-redundancy}~\cite{feng2021language,yuan2019scaffolds} pursuing a summary without redundancy,
and \textbf{factualness/consistency}~\cite{feng2020dialogue,zhao2021give,lei2021finer,kim2022mind} evaluating whether the summary is consistent with the source dialogue. There are also some typical fine-grained metrics evaluating errors in generated summaries mentioned in previous works~\cite{chen2020multi,chen2021dialsumm,liu2021coreference}: 
\textbf{Information missing} means that content mentioned in references are missing in generated summaries, while \textbf{information redundancy} is the opposite.
\textbf{Reference error} refers to wrong associations between a speaker and an action or a location.
\textbf{Reasoning error} is that the model incorrectly reasons the conclusion among multiple dialogue turns.
Moreover, \citet{chen2020multi} mentioned \textbf{improper gendered pronouns} referring to improper gendered pronouns. \citet{tang2021confit} also proposed \textbf{circumstantial error}, \textbf{negation error}, \textbf{object error}, \textbf{tense error} and \textbf{modality error} for more detailed scenarios. All of their error types can also be grouped into two classes, where the information missing and redundancy are for the coverage of key information, and the rest are for factual consistency.

A summary of evaluation metrics adopted in existing dialogue summarization works is in Table~\ref{tab:eval-metrics}.

\begin{table}[h]
	\centering
	\small
	\begin{tabular}{|l|l|p{6.8cm}|}
		\hline
		\textbf{Types} & \textbf{Description} & \textbf{Metrics} \\
		\hline
		\multirow{3}{*}{Automatic Evaluation} & Commonly-used & Rouge, MoverScore, BERTScore, BARTScore, ... \\
		\cline{2-3}
		 & Information Coverage & medical concept coverage, critical information completeness, factual completeness, ...\\
		 \cline{2-3}
		 & Factual Consistency & QA-based metrics, NLI-based metrics, binary classifiers with synthetic data, negation correctness, ...\\
		 \hline
		\multirow{3}{*}{Human Evaluation} & Evaluation Aspects& readability / fluency, informativeness, conciseness / non-redundancy, factualness / consistency\\
		\cline{2-3}
		& Error Types & {information missing, information redundancy, reference error, reasoning error, improper gendered pronouns, circumstantial error, negation error, object error, tense error, modality error} \\ 
		\hline
		
	\end{tabular}
	\caption{A summary of evaluation metrics.}
	\label{tab:eval-metrics}
\end{table}




\section{Analysis and Future Directions}
In this section, we first present a statistical analysis of the papers 
covered in this survey. Then, some future directions are proposed inspired by
our observations.
\begin{figure}[htbp]
	\centering
	\begin{minipage}[t]{0.45\linewidth}
		\centering
		\includegraphics[scale=0.6]{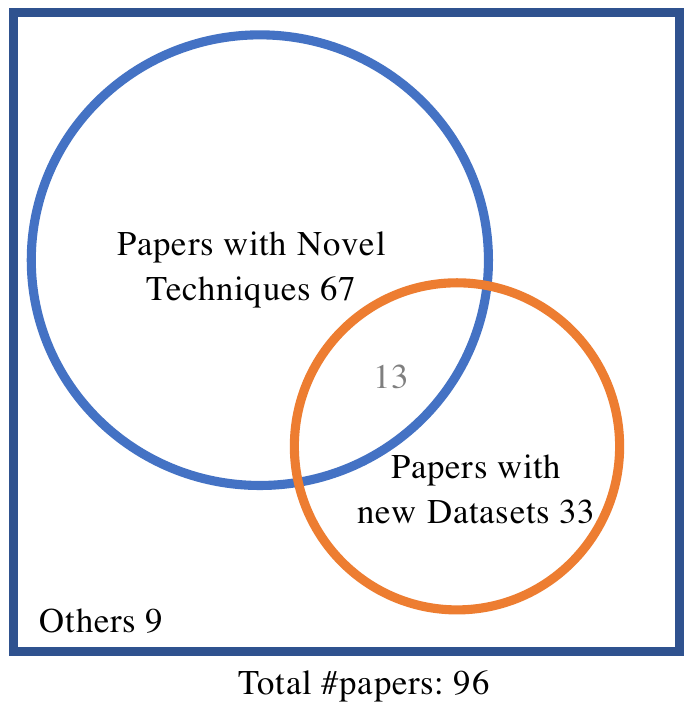}
		\caption{Statistics of abstractive dialogue \\summarization papers.}
		\label{fig:papers}
	\end{minipage}
	\begin{minipage}[t]{0.45\linewidth}
		\centering
		\includegraphics[scale=0.6]{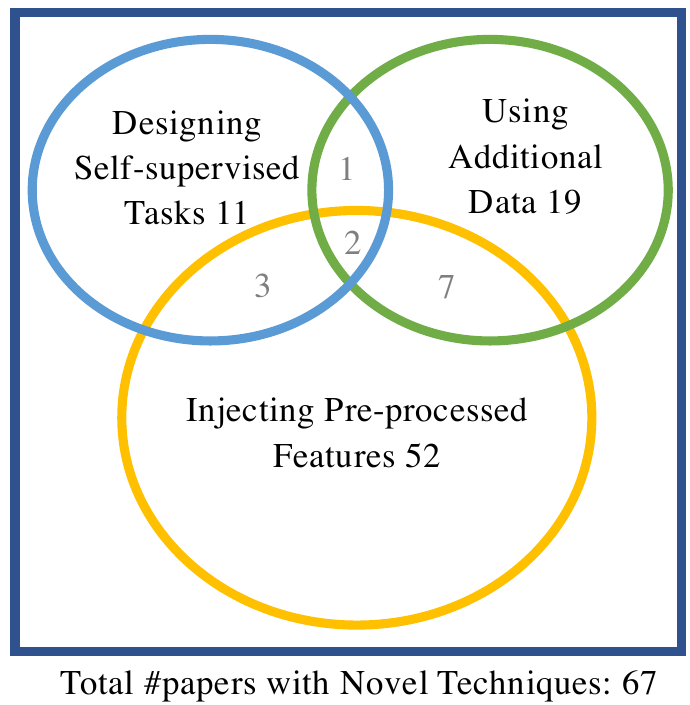}
		\caption{Statistics of papers with technical contributions.}
		\label{fig:technical}
	\end{minipage}
\end{figure}
\subsection{Paper Analysis}\label{sec:observations}
\begin{figure}[ht]
	\centering
	\subfigure[]{
	\begin{minipage}[t]{0.3\linewidth}
	\centering
	\includegraphics[scale=0.4]{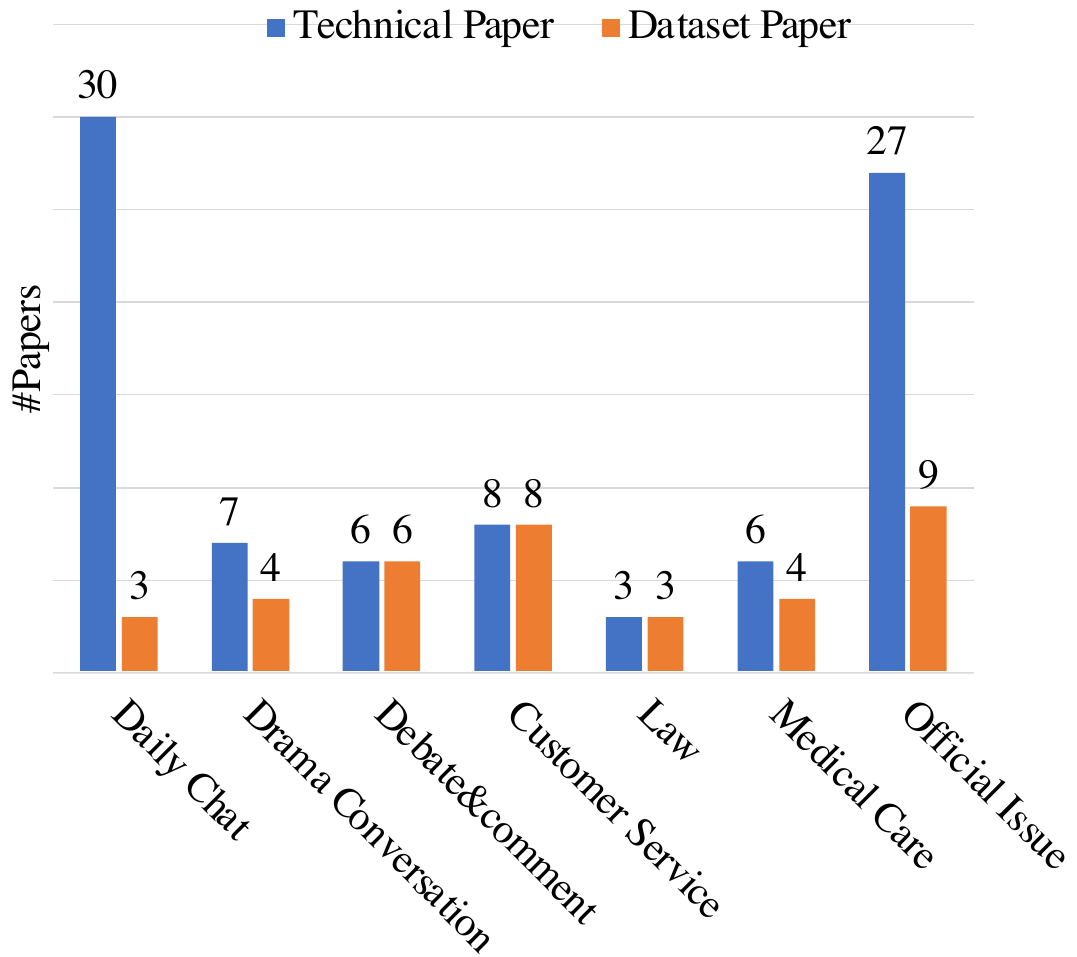}
	\label{fig:tech-data}
	\end{minipage}}
	\subfigure[]{
	\begin{minipage}[t]{0.3\linewidth}
	\centering
	\includegraphics[scale=0.4]{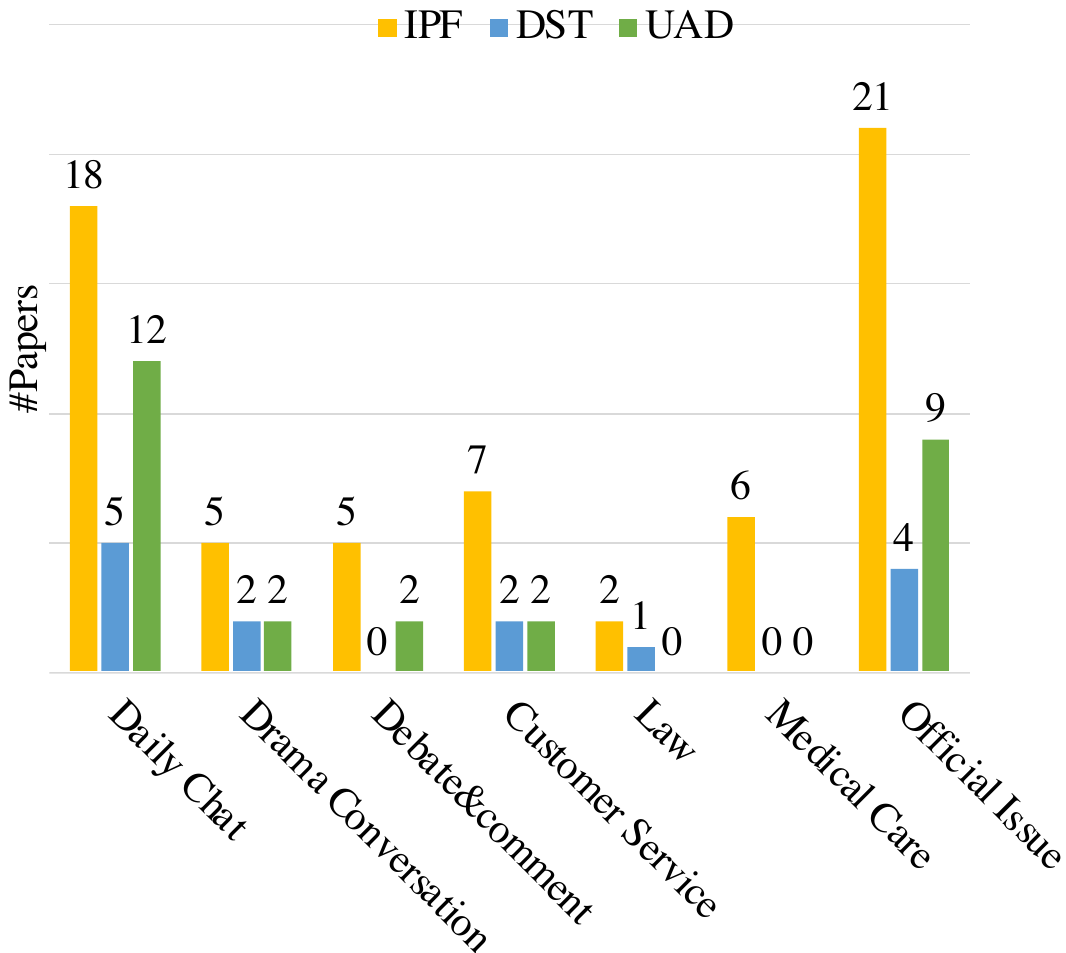}
	\label{fig:tech-scenario}
	\end{minipage}}
	\subfigure[]{
	\begin{minipage}[t]{0.3\linewidth}
	\centering
\includegraphics[scale=0.38]{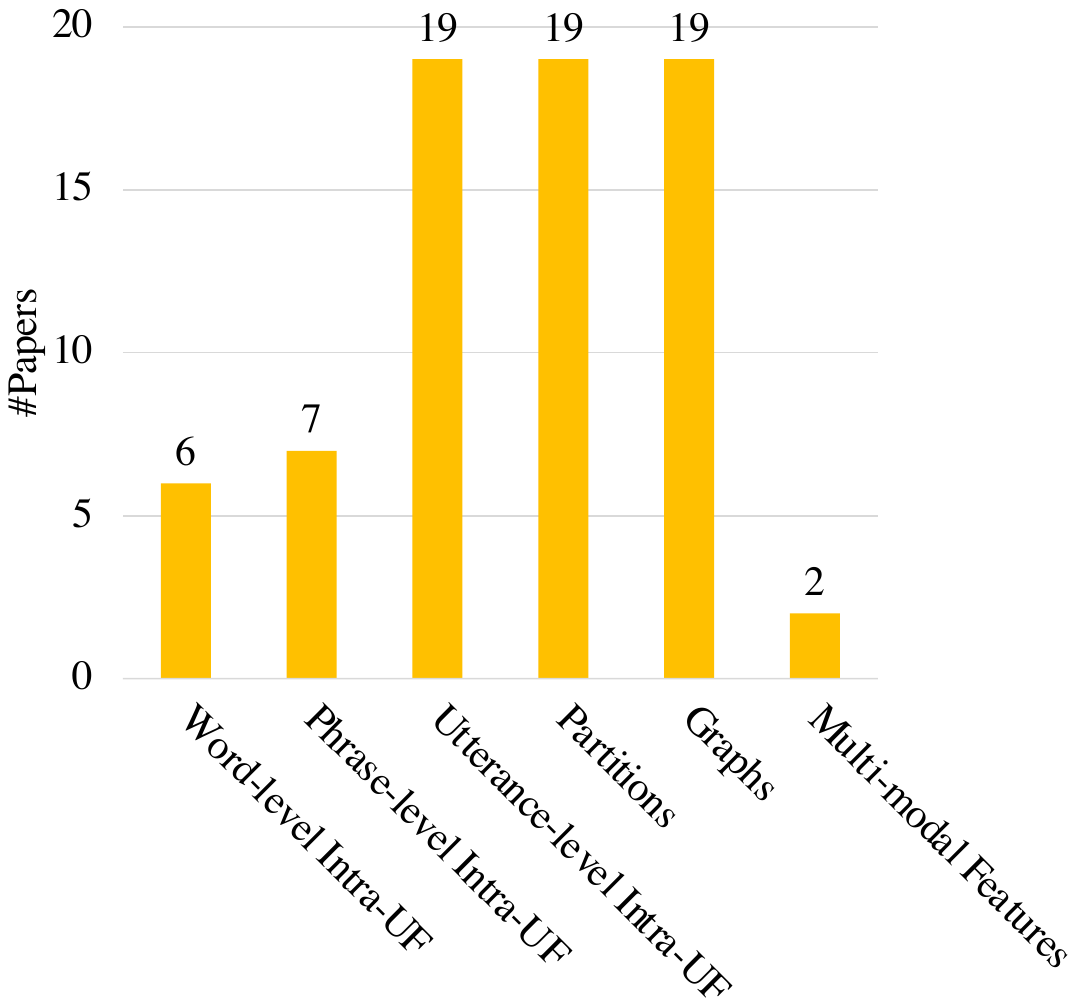}
\label{fig:feature-scenario}
\end{minipage}}
\caption{(a) The number of technical papers and dataset papers under different scenarios. (b) The number of technical papers dividing by directions under different scenarios. IPF, DST and UAD are short for the three directions. (c) The number of technical papers under different features. Intra-UF is intra-utterance features.}
\end{figure}

The total number of papers on abstractive dialogue summarization investigated 
in this survey is $96$. 
As shown in \figref{fig:papers}, $33$ of them propose new datasets and 
$67$ make novel technical contributions. The other $9$ papers are either 
a survey, a demo, or other strongly related papers. 
The overall ratio between technical papers and dataset papers 
(\textit{tech-data ratio}) is around $2.03:1$. Compared with the number of 
papers under different application scenarios in \figref{fig:tech-data}, 
we found that scenarios of daily chat and official issues receive more 
attentions, which is evident from the larger number of papers and 
greater tech-data ratios of more than $3$.
However, the other scenarios are less explored, with much lower 
tech-data ratios ranging from $1.0$ to $1.75$.
There is no significant difference in the number of datasets between
well-researched domains and the others. However, the release time and availability of different datasets vary.
AMI and ICSI are well-known meeting summarization datasets released in the early stage of the $20$th century, while most other datasets have been proposed in recent years.
Datasets for daily chat are all publicly available, while datasets for medical care and laws are not accessible to the majority of researchers. It's a good sign that high-quality corpora, such as AMI and SAMSum, lead to a prosperous of techniques for dialogue summarization, but also raise a worry about the generalization ability of current techniques because of their over-reliance on specific datasets which may lead to over-fitting.


The distribution of technical papers in each of the three research directions 
is shown in \figref{fig:technical}.
While $11$ and $19$ papers focus on designing self-supervised tasks and 
using additional data, respectively, 
more than $77\%$ of the entire body of works targets the injection of 
pre-processed features. The trends of paper account for different techniques 
across scenarios that are similar to each other according to the statistics 
in \figref{fig:tech-scenario}. IPF, DST, and UAD are short for injecting pre-processed features, designing self-supervised tasks, and using additional data, respectively.
The number of papers using features under different categories is 
shown in Figure~\ref{fig:feature-scenario}. 
And based on these $52$ paper, we go for a deep insight into 
correlations between features and applications scenarios by categorizing 
papers according to features and their tested scenarios in 
Table~\ref{tab:correlation}. 

\begin{table}
	\centering
	\begin{tabular}{|l|ccc|cc|c|}
		\toprule[1pt]
		\multicolumn{1}{|c|}{\multirow{2}{*}{\diagbox{Scenarios}{Features}}} & \multicolumn{3}{c|}{\textbf{Intra-Utterance Features}} &\multicolumn{2}{c|}{\textbf{Inter-Utterance Features}} & \multirow{2}{*}{\textbf{\makecell{Multi-\\modal\\ Features}}} \\
		
		& \makecell[c]{Word\\level} & \makecell[c]{Phrase\\level} & \makecell[c]{Utterance\\level} & Partitions & Graphs& \\
		
		\midrule[1pt]
		\multicolumn{7}{|l|}{\textit{Open-domain Dialogue Sumamrization}}\\
		\hline
		
		\makecell{Daily\\Chat}
		&\makecell{\cite{prodan2021prompt}} 
		&\makecell{\cite{feng2021language}\cite{khalifa2021bag}\\\cite{park2022unsupervised}\cite{wu2021controllable}}
		&\makecell{\cite{asi2022end}\cite{feng2021language}\\\cite{kim2022mind}\cite{lei2021hierarchical}\\\cite{lei2021finer}\cite{prodan2021prompt}\\\cite{wu2021controllable}\cite{zechner2002automatic}} 
		&\makecell{\cite{asi2022end}\cite{chen2020multi}\\\cite{feng2021language}\cite{liu2021topic}} 
		& \makecell{\cite{chen2021structure}\cite{feng2021incorporating}\\\cite{lei2021finer}\cite{liu2021controllable}\\\cite{liu2021coreference}\cite{park2022unsupervised}\\\cite{zhao2021give}\cite{liu2023picking}}
		&\makecell{-}\\
		
		\hline
		\makecell{Drama\\ Conversation}
		&\makecell{-} 
		&\makecell{\cite{park2022unsupervised}}  
		&\makecell{-}  
		&\makecell{\cite{li2021hierarchical}\cite{liu2021topic}\cite{zhang2021summ}} 
		& \makecell{\cite{park2022unsupervised}\cite{zhao2021give}} 
		&\makecell{-}  \\
		
		\hline
		\makecell{Debate \& Comment}
		&\makecell{-} 
		& \makecell{\cite{park2022unsupervised}} 
		&\makecell{\cite{yang2022tanet}}  
		&\makecell{-}  
		&\makecell{\cite{chen2021structure}\cite{fabbri2021convosumm}\\\cite{feng2021incorporating}\cite{park2022unsupervised}\cite{yang2022tanet}} &\makecell{-}  \\
		
		\midrule[1pt]		 
		
		\multicolumn{7}{|l|}{\textit{Task-oriented Dialogue Sumamrization}}\\
		\hline
		
		\makecell{Customer Service}
		& \makecell{-}
		&\makecell{\cite{zou2021topic}} 
		& \makecell{\cite{asi2022end}\cite{yang2022tanet}\\\cite{yuan2019scaffolds}\cite{zhang2020unsupervised}\\\cite{zou2021topic}}
		& \makecell{\cite{asi2022end}\cite{zou2021unsupervised}}
		&  \makecell{\cite{yang2022tanet}\cite{yuan2019scaffolds}\\\cite{zhao2021todsum}}
		&  \makecell{-} \\
		
		\hline
		\makecell{Law}
		& \makecell{-} 
		& \makecell{\cite{gan2021inspectional}} 
		&\makecell{\cite{duan2019legal}\cite{gan2021inspectional}} 
		& \makecell{-}
		&  \makecell{-}
		& \makecell{-} \\
		
		\hline
		\makecell{Medical Care}
		&\makecell{-} 
		&  \makecell{\cite{joshi2020dr}}
		&\makecell{\cite{song2020summarizing}} 
		&\makecell{\cite{krishna2021generating}\cite{liu2019topic}\cite{zhang2021leveraging}} 
		&  \makecell{\cite{molennar2020healthcare}}
		& \makecell{-}\\
		
		\hline
		\makecell{Official\\ Issue\\(Meeting\&Email)}
		&\makecell{\cite{murray2005extractive}\cite{OyaMCN14}\\\cite{qi2021improving}\cite{singla2017spoken}\\\cite{zhu2020end}} 
		&\makecell{\cite{feng2021language}\cite{park2022unsupervised}} 
		&\makecell{\cite{di2020da}\cite{feng2021language}\\\cite{goo2018abstractive}\cite{murray2005extractive}\\\cite{qi2021improving}\cite{yang2022tanet}\\\cite{zhu2020end}} 
		&\makecell{\cite{banerjee2015generating}\cite{di2020da}\cite{feng2021language}\\\cite{koay2021sliding}\cite{li2019keep}\cite{liu2021dynamic}\\\cite{qi2021improving}\cite{shang2018unsupervised}\\\cite{zhang2021summ}\cite{zheng2020abstractive}\\\cite{zhong2021qmsum}} 
		&\makecell{\cite{banerjee2015generating}\cite{feng2020dialogue}\cite{ganesh2019restructuring}\\\cite{MehdadCTN13}\cite{OyaMCN14}\\\cite{park2022unsupervised}\cite{shang2018unsupervised}\\\cite{yang2022tanet}} 
		&\makecell{\cite{li2019keep}\cite{murray2005extractive}}\\
		
		\bottomrule[1pt]

	\end{tabular}
	\caption{Existing work on injecting pre-processed features for 
different scenarios. The taxonomy of features and dialogue summarization 
scenarios are in the columns and rows respectively. The same work may appear 
multiple times in the table since it might have experimented with multiple 
datasets under various scenarios and utilized features in different groups.}
	\label{tab:correlation}
\end{table}

We make following observations:
\begin{itemize}
	\item Scenarios of Official Issue and Daily Chat attracted the most 
	attentions while other scenarios lack research as mentioned before. 
	
		\item Utterance-level intra-utterance features and inter-utterance 
	features are widely exploited, indicating that modeling utterance-level or 
	beyond utterance-level features is more effective at contextual dialogue 
	understanding. Among them, speaker/role information and topic transitions are 
	two main common features which work well 
	under both ODS and TDS scenarios. There is also a lack of attention on 
	multi-modal features: only two papers have investigated it, possibly due to the scarcity of multi-modal datasets.
		\item Word-level and phrase-level intra-utterance features are no longer
	required with the wide adoption of pre-trained language models, except in
	integrating domain dictionaries in TDS. These features, especially keywords,
	are preferred to use as nodes for further constructing graphs, which helps 
	capture the global information flows for both ODS and TDS. 
		\item Partitions are extremely effective for TDS where dialogues 
	are usually long with inherent semantic transitions, such as agendas for meetings and domain shifts in customer service. 
	Identifying these transitions achieves a high degree of consensus among annotators. 
	In contrast, semantic flows in ODS are often interleaved in a complex fashion,
	which can be better represented as graphs, such as discourse graphs and topic graphs.

\end{itemize}

\subsection{Future Directions}
\label{sec:future}

We discuss some possible future directions and organize them into
three dimensions: \textit{task scenarios}, \textit{approaches} and \textit{evaluations}. 

\subsubsection{More Complicated and Controllable Scenarios}

More newly explored scenarios such as multilingual dialogue summarization, multi-modal dialogue summarization, multi-session dialogue summarization are worth researching. In addition, we put forward {personalized dialogue summarization} as a novel future direction that pays special attention to speaker or reader-related information neglected by previous works.

\textbf{Multilingual dialogue summarization} is a rising topic with few related papers. It considers multiple languages existing in the dialogue and summary on three levels of granularity. First, the most fine-grained one considers interactions between peers who are fluent in multiple languages resulting in the intra-utterance multilingual phenomenon is called ``code-mixing'' strictly~\cite{mehnaz2021gupshup}. Second, dialogues happening among multinational participants where they use their mother tongue to communicate lead to the inter-utterance multilingual phenomenon is called ``code-switching''~\cite{mehnaz2021gupshup}, i.e., mix-lingual in~\cite{feng2022msamsum}'s work. Third, summarizing a monolingual dialogue in a different language is called ``cross-lingual'' in~\citet{wang2022clidsum}. Different multilingual summarization datasets have been constructed for these settings based on the SAMSum~\cite{gliwa2019samsum}, DialogSum~\cite{chen2021dialsumm}, MediaSum~\cite{zhu2021mediasum} and QMSum~\cite{zhong2021qmsum} by either human annotations~\cite{wang2022clidsum,mehnaz2021gupshup,chen2022cross} or automatic machine translation~\cite{feng2022msamsum}. Preliminary studies in these papers show the potential of end-to-end multilingual models, such as mBART~\cite{tang2021multilingual},  in this task and their weaknesses in low-resource languages, poor domain transfer ability among datasets~\cite{wang2022clidsum} and decreases in the performance when processing multiple languages with a single model~\cite{feng2022msamsum}. \citet{chen2022cross} proposed the cross-lingual conversation summarization challenge, paving the way for the prosperity of research in this direction. Our survey focuses on the taxonomy of approaches for monolingual dialogue summarization, which we expect to provide a backbone for this raising area.


\textbf{Multi-modal dialogue summarization} refers to dialogues occurring in 
multi-modal settings, which are rich in non-verbal information 
that often complements the verbal part and therefore contributes to 
summary contents. Some early work did research on speech dialogue summarization. 
However, most of them only extract audio features from speech and text 
features from ASR transcripts independently to produce extractive summaries. 
There is also work on video summarization~\cite{hussain2021video} focusing 
on highlighting critical clips while a textual summary is not considered.
Fusing the synchronous and asynchronous information among modalities 
is still challenging. AMI and ICSI are still valuable resources for 
research on multi-modal dialogue summarization.

\textbf{Multi-session dialogue summarization} is required when conversations 
occur multiple times among the same group of speakers. 
The Information mentioned in previous sessions becomes their consensus and 
may not be explained again in the current session. 
The summary generated merely from the current session is unable to 
recover such information and may lead to implausible reasoning. 
A similar multi-session task has been proposed by~\citet{xu2021beyond}. 
This setting also has some correlations with life-long learning~\cite{shuster2020deploying,liu2021lifelong}.
Such multi-session dialogues are just open-domain dialogues in drama or TV shows, such as SubTitles~\cite{malykh2020sumtitles} and SummScreen~\cite{chen2021summscreen}. However, current approaches generally break down the long dialogue and summary into shorter chunks due to the limitation of current models. 
For task-oriented scenarios, it is also common in real life. For example, the customer may repeatedly ask for help from the agent for the same issue that hasn't been solved before. An updated summary considering all of the questions and the latest answers can remind participants of the long dialogue history and can therefore facilitate the negotiation process. As far as we know, none of the papers have discussed this problem before in the area of dialogue summarization.





Recent work mainly focuses on summarizing the dialogue content but ignores the speaker-related or reader-related information. 
However, these are significant and negligible factors as reflected by diversified reference summaries by different annotators, both in the aspect of styles and the content selection.
Thus, we propose \textbf{personalized dialogue summarization}, which can be understood in two ways. 
On the one hand, a personalized dialogue refers to the consideration of 
personas for interlocutors in dialogues. For example, the character 
role-playing information is indispensable information for generating 
summaries given dialogue from CRD3~\cite{rameshkumar2020storytelling} in drama conversation scenarios.
On the other hand, it refers to generating different dialogue summaries for 
different readers or speakers. \citet{tepper2018personal} is a demo paper raising the 
requirements for personalized chat summarization. They did the first trial 
on this task considering the personalized topics of interests and social ties 
during the selection of dialogue segments to be summarized. 
Some task-oriented datasets, such as CSDS~\cite{lin2021csds}, contain summaries from both the user and agent aspect are similar
to problem here. Recent work from~\citet{lin2022other} solved this problem by adding the cross attention interaction and the decoder self-attention interaction to interactively acquire other roles' critical information. This work is designed only for scenarios with two roles.
Open-domain scenarios pose more challenges with a variety number of speakers and summary readers from different social groups, raising an
expectation for corresponding datasets and approaches, which is a possible interdisciplinary research orientation.

\subsubsection{Innovations in Approach} 
Approach innovations include four parts:  feature analysis, person-related features, generalizable and non-labored techniques, and the robustness of models.




From \secref{sec:observations}, we notice that although tens of papers 
introduce different features for dialogue summarization, there is still 
a lot of work to do. Comprehensive experiments to \textbf{compare the} \textbf{features} and 
their combinations upon the same benchmark are still needed, 
for features both in the same category or across categories. 
One can consider unifying the definition of similar features, 
such as different classification criteria of discourse relations or different graphs emphasizing the phrase-level semantic flows. 
These analyses would help in designing features in new applications and interpreting dialogue models.

More \textbf{person-related features} can be introduced to this task, 
such as speaker personalities~\cite{zhang2019consistent} and 
emotions~\cite{majumder2019dialoguernn}. Knowing the background of a speaker can help better understand the motivation behind each utterance and potentially influential in the selection of content to be summarized, especially for personalized dialogue summarization.
A plug-and-play mechanism on top of the decoder for persona-controlled summarization may be an ideal solution inspired by~\citet{DathathriMLHFMY20}.

\textbf{Generalizable and non-labored techniques} have attracted increasing attention on other dialogue modeling tasks, 
such as multi-turn response selection~\cite{xu2021learning} and 
dialogue generation~\cite{zhang2019consistent}. 
These works proposed different self-supervised training tasks, largely releasing the requirement on human labor.
From the distribution of technical papers for dialogue summarization in Figure~\ref{fig:technical}, we can see that approaches overwhelmingly 
rely on injecting pre-processed features. 
However, most of these approaches are labor-intensive since training a labeling tool requires human annotations of corresponding labels on some in-domain training data or needs trial and error to find the best hyper-parameters for transferring to the target summarization domain. Otherwise, it will suffer from error propagation and lead to poor performance.
Complicated features such as graphs summarized in Section~\ref{sec:graphs} tend to overfit specific domains or current datasets by human observations,
which have poor generalization ability and lead to tiring feature engineering works. More work exploiting the common nature among scenarios and exploring useful representations with language models is expected. 
Recently, large language models~(LLMs) with tons of billions of parameters like GPT-3~\cite{brown2020language} and LLaMA~\cite{touvron2023llama} have demonstrated drastically lifted text generation ability compared to previous pre-trained language models. Specifically, the decoder-only Transformer architecture is adopted as the fundamental backbone for almost all performant LLMs. Scaling decoder-only language models not only simplifies the architectural design dimensions but also enables a unified human-machine interface for various downstream language tasks. 
To accomplish summarization task, LLMs are typically prompted with instructions like ``\textit{Summarize the above article:}'' or chain-of-thought~\cite{cot,wang-etal-2023-element} methods that elicit LLMs to extract various features, e.g., entities, events, that are helpful to compose the final summary. Compared with traditional dialogue summarization systems, LLM-based methods largely alleviate the tedious human labor and can be more generalizable due to the removal of unintended annotation artifacts. Nevertheless, approaches that are previously applied to small pre-trained language models in this survey may also provide inspirations and be adapted to augment LLMs for better dialogue summarization performance.

Approaches nowadays are mostly built on the pre-trained language models, which are sensitive to trivial changes on other tasks~\cite{wang-etal-2022-rely,yan2022robustness}. 
Nevertheless, the \textbf{robustness of models} hasn't been widely-investigated in dialogue summarization. The only work from~\citet{jia2023reducing} proposes that switching an un-grounded speaker name shouldn't influence the models' generation. However, according to their experiments with BART fine-tuned on SAMSum, such changes can lead to dramatically different summaries with information divergence and various reasoning results and show over 14\% changes in Rouge scores. This may further result in unintended ethical issues by showing discrimination against specific groups of names. Thus, analysis of models' robustness and developing more insensitivity approaches are in urgent need for practical applications.

\subsubsection{Datasets and Evaluation Metrics}
Expectations on datasets and evaluation metrics for dialogue summarization are as follows.

Section~\ref{sec:observations} shows that \textbf{high-quality datasets} 
expedite the research. Besides the expectations on benchmark datasets for the above emerging scenarios, datasets for task-oriented dialogue summarization 
with privacy issues are also sought after. They can be in small sizes with 
real cases after anonymization or can be collected by selecting 
drama conversations in specific scenarios and annotated with 
domain experts. 

\textbf{Evaluation metrics} are significant which guides the improvement directions for upcoming models. However, 
widely used evaluation metrics in Sec.~\ref{sec:evalmetric} are all borrowed from document summarization 
tasks and their effectiveness is unverified. 
Recent work from~\citet{gao2022dialsummeval} re-evaluated 18 evaluation metrics and did a unified human evaluation with 14 dialogue summarization models on SAMSum dataset. 
Their results not only show the inconsistent performances of metrics between document summarization and dialogue summarization, and none of them excel in all dimensions for dialogue summarization, but also raise a warning on rethinking whether recently proposed complex models and fancy techniques truly improve the basic pre-trained language model.
Considering that human evaluation results are difficult to reproduce due to variations of annotator background and unpredictable situations in the annotation progress~\cite{clark2021all},
automatic metrics specially designed for dialogue summarization are 
urgently needed.

Factual errors caused by the mismatch between speakers and events are 
common as a result of complicated discourse relations among utterances 
in dialogues. 
Previous work~\cite{huang2021factual} on document summarization classifies 
factual errors into two types. One is intrinsic errors, referring to the fact 
contradicting the source document. The other is extrinsic errors, 
referring to unrelated facts.  
This classification is also suitable for dialogue summarization. However, whether their proposed 
QA-based~\cite{wang2020asking} and NLI-based~\cite{falke2019ranking} 
automatic evaluation approaches can be directly transferred to 
dialogue summarization for comparisons between dialogues and 
generated summaries due to their format disparity is still doubtful without thoughtful evaluations.
\citet{tang2021confit} introduced a taxonomy of factual errors for abstractive summarization and did human evaluation based on this categorization without proposing new automatic metrics.
\citet{liu2021controllable} made the first attempt by inputting the 
dialogue and summary together into a BERT-based classifier and claimed 
high accuracy on their own held-out data. But there is still a lack of 
details and comparisons to other methods, such as using bi-encoder 
architectures for the dialogue and summary respectively.
\citet{wang2022analyzing} classified factual errors in a similar way to ~\citet{tang2021confit} and propose a model-level evaluation schema for discriminating better summarization models, which is different from the widely-accepted sample-level evaluation schema that scores generated summaries and can further scoring the model based on their corresponding outputs. They evaluated the model by calculating the generation probability of faithful and unfaithful summaries collected by rule-based transformations based on their taxonomy. The generalization ability for this work among different datasets and scenarios is doubtful, since a similar work for news summarization, FactCC~\cite{kryscinski2020evaluating}, which is a metric trained based on rule-based synthetic datasets shows a poor generalization ability by~\citet{laban2022summac}.
With the strong generation ability of current LLMs, there's also a doubt that whether the previous taxonomy of error types and evaluation metrics is still suitable.
In a word, both \textbf{meta-evaluation benchmarks} and \textbf{evaluation methods} call for innovations.


\section{Conclusion}

Dialogue summarization is receiving increasing demands in recent years for releasing the burden of manual summarization and achieving efficient dialogue information digestion.
It is a cross-research direction of dialogue understanding and summarization.
Abstractive text summarization is a natural choice for dialogue 
summarization due to the characteristics of dialogues, including information 
sparsity, context-dependency, and the format discrepancy between 
utterances in first person and the summary from the third point of view. 
With the success of neural-based models especially pre-trained language models, 
the quality of generated abstractive dialogue summaries appears to be promising
for real applications.
This survey summarizes a wide range of papers on the subject.
In particular, it presents a hierarchical taxonomy for task scenarios, made up of two broad categories, i.e., open-domain dialogue summarization and task-oriented dialogue summarization.
A great many techniques developed in different approaches are categorized into three directions, including injecting pre-processed features, designing self-supervised tasks, and using additional data.
We also collect a number of evaluation benchmarks proposed so far and provide a deep analysis with valuable future directions.
This survey is a comprehensive checkpoint of dialogue summarization research thus far
and should inspire the researchers to rethink this task and 
search for new opportunities, especially with current LLMs. It is also a useful guide for engineers 
looking for practical solutions.

\bibliographystyle{ACM-Reference-Format}
\bibliography{csur}

\end{document}